\documentclass[10pt,twocolumn,letterpaper]{article}

\usepackage{wacv}
\usepackage{times}
\usepackage{epsfig}
\usepackage{graphicx}
\usepackage{amsmath}
\usepackage{amssymb}
\usepackage{booktabs}
\usepackage{multirow}
\usepackage{booktabs}

\newcommand*{\affaddr}[1]{#1} 
\newcommand*{\affmark}[1][*]{\textsuperscript{#1}}
\newcommand*{\email}[1]{\texttt{\small#1}}
\usepackage{xcolor}
\definecolor{dgreen}{rgb}{0.,0.6,0.}
\usepackage{caption}
\usepackage{subcaption}

\def\wacvPaperID{172} 

\wacvapplicationstrack 

\wacvfinalcopy 

\ifwacvfinal
\usepackage[breaklinks=true,bookmarks=false]{hyperref}
\else
\usepackage[pagebackref=true,breaklinks=true,colorlinks,bookmarks=false]{hyperref}
\fi

\begin{document}

\title{MFFN: Multi-view Feature Fusion Network for Camouflaged Object Detection}


\author{
Dehua Zheng\affmark[1]\quad Xiaochen Zheng\affmark[2]\quad Laurence T. Yang\affmark[1]\thanks{Corresponding author.}\quad Yuan Gao\affmark[3]\quad Chenlu Zhu\affmark[4]\quad Yiheng Ruan\affmark[4]\\
\affaddr{\affmark[1]Huazhong University of Science and Technology, China\quad \affmark[2]ETH Zürich, Switzerland\\ \affmark[3]Hainan University, China\quad\affmark[4]Hubei Chutian Expressway Digital Technology, China}\\
\email{dwardzheng@hust.edu.cn\quad xzheng@student.ethz.ch\quad ltyang@gmail.com}
}
\maketitle
\thispagestyle{empty}

\begin{abstract}
\label{abs}
Recent research about camouflaged object detection~(COD) aims to segment highly concealed objects hidden in complex surroundings. The tiny, fuzzy camouflaged objects result in visually indistinguishable properties. However, current single-view COD detectors are sensitive to background distractors. Therefore, blurred boundaries and variable shapes of the camouflaged objects are challenging to be fully captured with a single-view detector. To overcome these obstacles, we propose a behavior-inspired framework, called \textbf{M}ulti-view \textbf{F}eature \textbf{F}usion \textbf{N}etwork~(MFFN), which mimics the human behaviors of finding indistinct objects in images, i.e., observing from multiple angles, distances, perspectives. Specifically, the key idea behind it is to generate multiple ways of observation (\textbf{multi-view}) by data augmentation and apply them as inputs. MFFN captures critical boundary and semantic information by comparing and fusing extracted multi-view features. In addition, our MFFN exploits the dependence and interaction between views and channels. Specifically, our methods leverage the complementary information between different views through a two-stage attention module called \textbf{C}o-\textbf{a}ttention of \textbf{M}ulti-\textbf{v}iew~(CAMV). And we design a local-overall module called \textbf{C}hannel \textbf{F}usion \textbf{U}nit~(CFU) to explore the channel-wise contextual clues of diverse feature maps in an iterative manner. The experiment results show that our method performs favorably against existing state-of-the-art methods via training with the same data. The code will be available at \url{https://github.com/dwardzheng/MFFN\_COD}.
\end{abstract}
\begin{figure}[htbp]  
\centering\includegraphics[width=8cm]{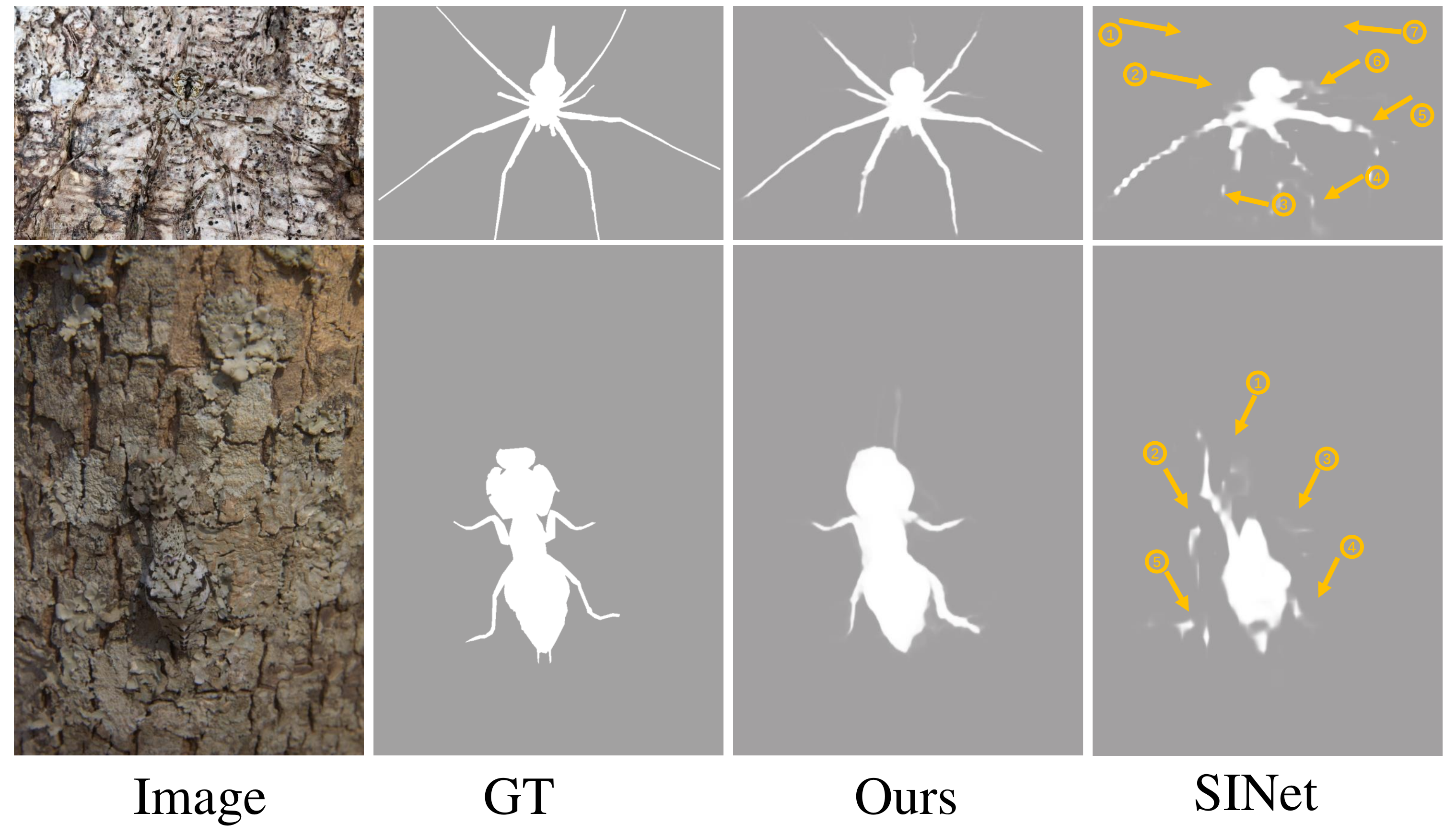} 
\caption{Visualization of camouflaged animal detection. The state-of-the-art and classic single-view COD model SINet~\cite{cod} is confused by the background sharing highly similarities with target objects and missed a lot of boundary and region shape information (indicated by orange arrows). Our multi-view scheme will eliminate these distractors and perform more efficiently and effectively.}
\label{fig:cct}   
\end{figure}

\section{Introduction}
\label{intro}
Camouflage is a mechanism~\cite{Camouflage_mechanisms} by which organisms protect themselves in nature. Camouflaged object detection~(COD) is a countermeasure against the camouflage mechanism, aiming to capture the slight differences between the object and the background to obtain accurate detection results. Unlike general object detection and salient object detection, in which the objects and background can be easily distinguished by human eyes or advanced deep learning models, COD is more challenging 
because it requires a sufficient amount of visual input and prior knowledge~\cite{cvp} to address the complicated problem caused by the highly intrinsic similarity between the target object and the background. Thus, COD has a wide range of valuable applications in promoting the search and detection of biological species~\cite{Camouflage}, assisting the medical diagnosis with medical images~\cite{UNet3, DLT}, and improving the detection of pests and diseases in agriculture~\cite{RDL}.

 Recently, many researches put emphasis on learning from a fixed single view with either auxiliary tasks~\cite{CAMO,rank_net,context_reasoning,MGL, BGSAN, ji2022deep}, uncertainty discovery~\cite{uncertainty_joint,aleatoric_uncertainty}, or vision transformers~\cite{uncertainty_transformer,pei2022osformer} and their proposed methods achieved significant progress. Nevertheless, due to visual insignificance of camouflaged objects and contextual insufficiency from single-view input, they are still striving to precisely recognize camouflaged objects and their performance needs to be improved. We found that the current COD methods are easily distracted by negative factors from deceptive background/surroundings, as illustrated in Fig.~\ref{fig:cct}. As a result, it is hard to mine discriminative and fine-grained semantic cues of camouflaged objects, making accurately segment camouflaged objects from a confusing background and predict some uncertain regions incapable. Meanwhile, we learn that when people observe a concealed object in images, they usually adjust the viewing distance, change the viewing angle, and change the viewing position to find the target object more accurately. Inspired by it, we aim to design a simple yet efficient and effective strategy. The aforementioned considerations motivate us to consider the semantic and context exploration problem with \textbf{multi-view}.
 We argue that corresponding clues, correlations, and mutual constraints can be better obtained by utilizing information from different viewpoint of the scene (e.g., changing observation distances and angles) as complementary. Furthermore, we argue that carefully designing the encoded feature fusion modules can help the encoder learn accurate information corresponding to boundary and semantics. Taking these into mind, our research will focus on the following three aspects: \textit{(1) how to exploit the effects of different types of views on COD task, and the combination of multi-view features to achieve the best detection effect; (2) how to better fuse the features from multiple views based on correlation awareness and how to enhance the semantic expression ability of multi-view feature maps without increasing model complexity; (3) how to incrementally explore the potential context relationships of a multi-channel feature map}.

To solve our concerned pain points of COD task, we propose a \textbf{M}ulti-view \textbf{F}eature \textbf{F}usion \textbf{N}etwork~(MFFN) for the COD task to make up for the semantic deficiency of fixed view observation. First, we use the multi-view raw data, which are generated by different data augmentation, as the inputs of a backbone extractor with shared weights. We implement a ResNet model as the backbone extractor integrating the feature pyramid network~(FPN)~\cite{FPN} to focus on object information of different scales. In addition, we design a \textbf{C}o-\textbf{a}ttention of \textbf{M}ulti-\textbf{v}iew~(CAMV) module to integrate multi-view features and to explore the correlation between different view types. CAMV consists of two stages of attention operation. In the first stage, the inherent correlation and complementary analysis are mainly conducted for multiple viewing distances and angles to obtain the view features with a unified scale. In the second stage, the external constraint relations between viewing angles and distances are further leveraged to enhance feature maps' semantic expression. For the enhanced multi-view feature tensor, we design a \textbf{C}hannel \textbf{F}usion \textbf{U}nit~(CFU) to further exploit the correlation between contexts. In the CFU module, we first carry out up-down feature interaction between channel dimensions and then carry out progressive iteration on the overall features. CAMV is applied to observe the multi-view attention features of different size feature maps of FPN architecture. The CFU module contains the previous layer's information as each size's feature maps are eventually restored to their original size. Finally, the final prediction results are obtained by sigmoid operation. The prediction further benefits from UAL design.

Our contribution can be summarized as follows:
1)~We propose MFFN model to solve the challenging problems faced by single-view COD models. MFFN can capture complementary information acquired by different viewing angles and distances and discover the progressive connection between contexts.

2)~We design the CAMV module to mine the complementary relationships within and between different types of view features and enhance the semantic expression ability of multi-view feature tensors, and use the CFU module to conduct progressive context cue mining.

3)~Our model is tested on three datasets of CHAMELEON~\cite{chameleon}, COK10K~\cite{cod} and NC4K~\cite{rank_net}, and quantitative analysis is conducted on five general evaluation indicators of $S_m$~\cite{Sm}, $F_\beta^w$~\cite{Fbw}, $MAE$, $F_\beta$~\cite{Fb} and $E_m$~\cite{Em}, all of which achieved superior results.

\begin{figure*}
\begin{center}
\includegraphics[width=13cm]{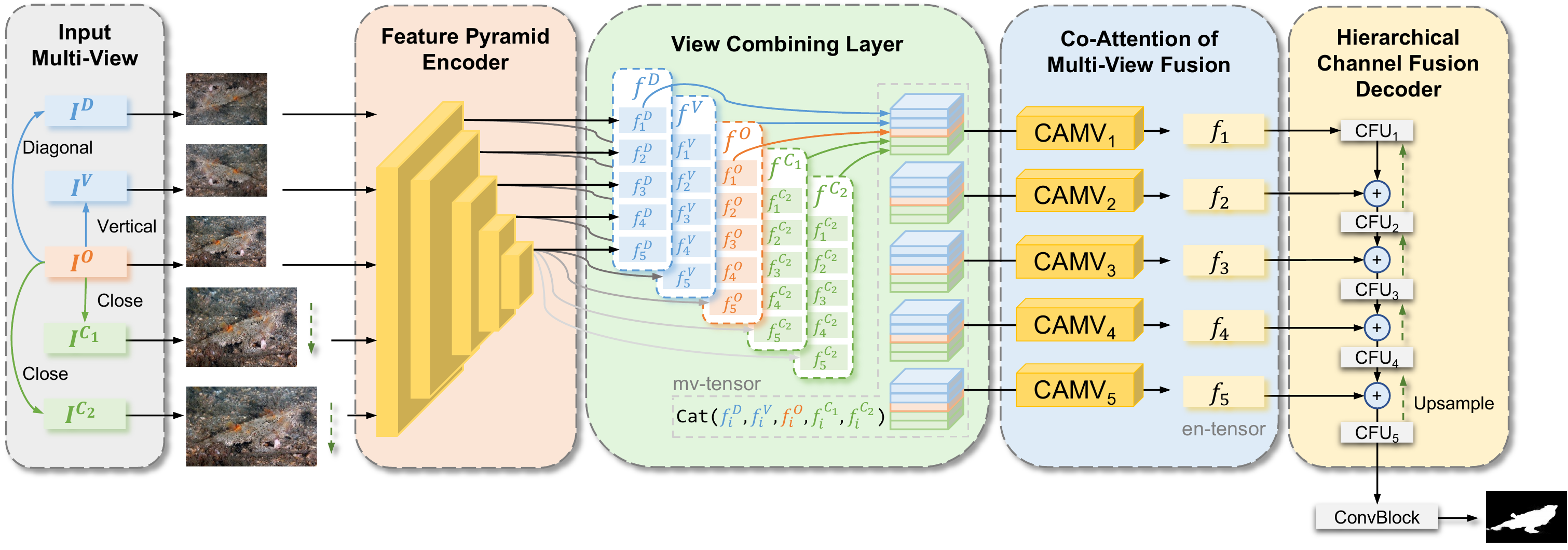}
\end{center}
   \caption{Overview of our model structure. We generate multiple views (\textbf{D}iagonally and \textbf{V}ertically flipped views, \textbf{C}lose looking views) of the data by different transformation methods. The shared pyramid feature encoder is applied to extract hierarchical features of different scales corresponding to different view choices. The view combining layer concatenates features of same level from different views ($f^D_i, f^V_i, f^O_i, f^{C_1}_i, f^{C_2}_i$) channel-wisely and output multi-view feature tensors (\textit{mv-tensor}s). The model feeds \textit{mv-tensor}s into CAMVs and obtain multi-view enhanced feature tensor (\textit{en-tensor}) $f_i$. CAMV is adopted to fuse features and aggregate vital clues between different views by a two-stage co-attention mechanism. The \textit{en-tensor}s are further decoded and the contextual correlation are exploited by hierarchical channel fusion unit simultaneously. In the end, a probability map of camouflaged object in the input image is computed by several convolutional blocks.}
\label{fig:overview}
\end{figure*}

\section{Related work}
\paragraph{Salient Object Detection~(SOD).} SOD is a kind of segmentation task in essence. It calculates saliency map first and then merges and segmented saliency object. In previous studies, traditional methods based on manual features pay more attention to color~\cite{color_1, color_2}, texture~\cite{texture_1, color_2}, contrast~\cite{contrast_1, contrast_2} and so on, but lack advantages in complex scenes and structured description. With the development of CNN, SOD algorithm has achieved leapfrog development. Li~\etal~\cite{MDF} combines local information with global information to overcome the problem of highlighting object boundary but not the overall object in the model based on local. The model structure design idea of multi-level features, has been more widely applied in~\cite{M_scal_1, M_scal_2, M_scal_3, M_scal_5}. Similar to COD, clear boundary information is crucial for SOD task~\cite{edge_1, edge_2, edge_3}. The development of attention mechanism provides more schemes for exploring the correlation between channel dimension and spatial dimension~\cite{attn_4, attn_5, attn_6}. The application of attention mechanism improves the performance of SOD model~\cite{attn_1, attn_2, attn_3}. SOD faces  simpler background surroundings. Although excellent performance can be obtained by applying relevant models to COD task, specific design is still needed to remove the interference from the background surroundings.

\paragraph{Camouflaged Object Detection (COD).} In recent years, some researches applied multi-task learning to detect the camouflaged objects. Le~\etal~\cite{CAMO} introduced the binary classification task as the second branch and auxiliary task of camouflaged object segmentation. Zhu~\etal~\cite{BGSAN} proposed a new boundary-guided separated attention network (BSA-NET), which uses two streams of separated attention modules to highlight the boundaries of camouflaged objects. Lv~\etal~\cite{rank_net} proposed a multi-task learning framework to jointly localize and segment the camouflaged objects while inferring their ranks. Zhai~\etal~\cite{MGL} designed a mutual graph learning model to detect the edge and region of the objects simultaneously. There are some uncertainty-aware methods. Li~\etal\cite{uncertainty_joint} proposed an uncertainty-aware framework containing a joint network for both salient and camouflaged object detection. Yang~\etal~\cite{uncertainty_transformer} introduced Bayesian learning into the uncertainty-guided transformer reasoning model. Liu~\etal\cite{aleatoric_uncertainty} designed an aleatoric uncertainty estimation network to indicate the prediction awareness. Sun~\etal~\cite{context-cross} placed emphasis on rich global context information with the integration of cross-level features. Pei~\etal~\cite{pei2022osformer} applied a one-stage location-sensing transformer and further fused the features from transformer and CNN. Some bio-inspired methods are proposed.  For example, \cite{ZoomNet-CVPR2022, context_reasoning, cod} use multi-scale information but from one single view. Meanwhile, \cite{ZoomNet-CVPR2022} shows single-view information is not sufficient for accurately detecting camouflaged objects. 
We hereby argue that view generation and selection might play an important role and we aim to develop our model by mimicking the behavior of humans when understanding complicated concealed objects by altering the way they observing an image. Our proposed method exploits the visual perception knowledge and semantic cues by aggregating complementary information from multi-view. Accordingly, our model is simple yet efficient and effective to comprehensively understand scene and to accurately segment the camouflaged objects.

\section{Method}

\paragraph{Motivation.} Motivated by the challenges of single-view COD models, we attempt to capture boundary and regional semantic information with rich viewing angles and flexible viewing distances. In order to merge diverse context information from features of multi-view inputs and FPN multi-level outputs, we design a feature fusion module based on two-stage attention mechanism to obtain enhanced feature tensors. It also avoids redundant structural design. To leverage the rich information contained in channel dimensions, we design a local-overall context/cues mining structure based on channel-wise integration. Meanwhile, it also enhances the information expression of the feature tensors.

\subsection{Multi-view Generation}
As shown in Fig.~\ref{fig:cct}, the single-view model misses necessary boundary, region, and shape information. Inspired by human behavior, taking complementary views of observation into account will overcome this defect and we design three different views: \textit{distance, angle, and perspective view}. We obtain different \textit{distance views} through the resize operation with the proportional interval of the resize operation larger than 0.5 to increase the distinction. We get different \textit{angle views} by mirror transformation, including horizontal, vertical and diagonal mirror transformation. We obtain different \textit{perspective views} through affine transformation. Specifically, three corresponding points on the original and the target image are selected as references to calculate the transformation matrix. The above operations are based on OpenCV and the implementation in OpenCV is in Appendix B. The ablation study proves that the combination of two angle views obtained by mirror transformation and two close distance views obtained by resize operation is an effective selection scheme. As shown in the Appendix A, our multi-view strategy can be easily transferred to the SOD task and achieve excellent performance in salient object detection (SOD) task.

\subsection{Architecture Overview}
The overview of our proposed MFFN is illustrated in Fig.~\ref{fig:overview}. MFFN adopts ResNet~\cite{resnet} as the backbone network for feature extraction, and adopts the FPN~\cite{FPN} to capture feature information of different levels from different views. We design the CAMV module to merge diverse context information and to capture complementary information from encoded multi-view features. Furthermore, we applied CFU module to fuse the channel-wise context information and clues in an iterative manner. As shown in Fig.~\ref{fig:overview}, given an input original image $I^O\in\mathbb{R}^{H\times W\times3}$, we create \textit{flipped} and \textit{close} views by applying mirror and resize transformation. The multi-view inputs are defined as \{$I^D\in\mathbb{R}^{H\times W\times3}$, $I^V\in\mathbb{R}^{H\times W\times3}$, $I^O\in\mathbb{R}^{H\times W\times3}$, $I^{C_1}\in\mathbb{R}^{H_1\times W_1\times3}$, $I^{C_2}\in\mathbb{R}^{H_2\times W_2\times3}$ \}, where $D, V$ indicate diagonally and vertically flipped views, $O$ indicates original view, and $C_1, C_2$ represent two different scale close views. We input each observed view into a backbone network with shared weights, and obtain feature maps of different levels through FPN~\cite{FPN}. Then, we apply CAMV module to fuse the multi-view feature tensors from a specific FPN stage by a two-stage attention mechanism. Furthermore, we design the CFU module to mine the contextual correlation and critical clues between the multi-view enhanced feature maps ${f_1, ..., f_5}$. Finally, MFFN restores the feature maps to its original size by gradual upsampling structure, so as to obtain the final output results.

\subsection{Co-attention of Multi-view}
The COD methods proposed in recent years pay more attention to feature mining from a fixed view and thus ignore information complemented from different views. Inspired by the biological mechanism, visual information from different ways of observing and watching can be correlated and complemented. Based on the above discoveries, we implement CAMV, consisting of a two-stage attention mechanism to complement boundary information with features from different viewing angules and enhance semantic information with different viewing distance. CAMV reduces redundant network design through multi-view interaction and fusion.

\begin{figure*}
\begin{center}
\includegraphics[width=13cm]{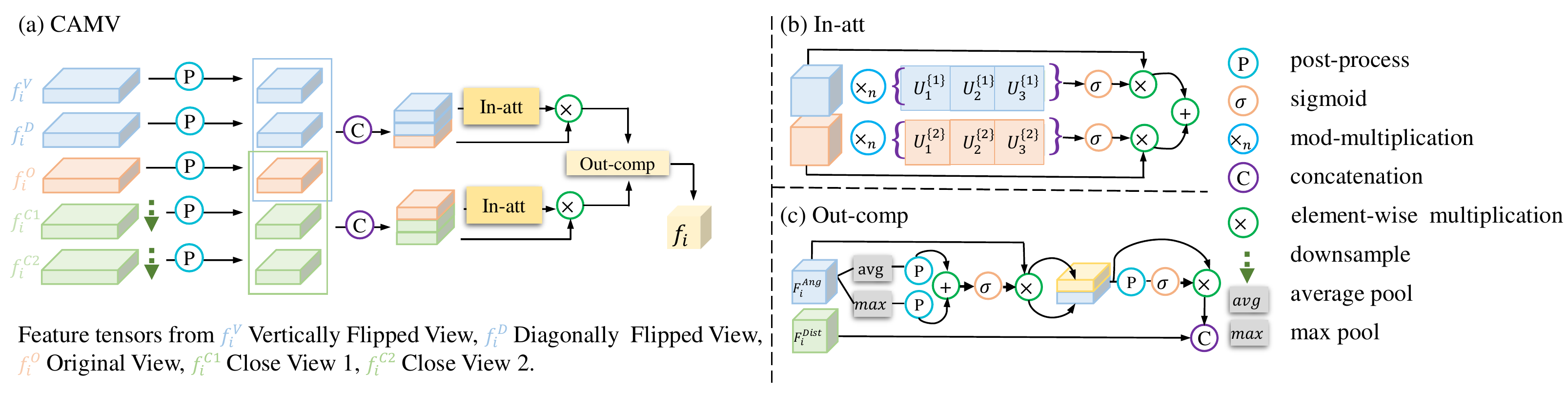}
\end{center}
   \caption{The architecture of our CAMV module. First, the two view types are processed by intra-class attention (\textit{In-att}) mechanism, and then the two view types are fused by the complementation of external classes (\textit{out-comp}). In a nutshell, CAMV consists of a two-stage attention mechanism.}
\label{fig:camv}
\end{figure*}

The framework of CAMV is shown in Fig.~\ref{fig:camv}. Since the scales of multiple viewing distances features $f_i^{C_1}\in\mathbb{R}^{h_1\times w_1\times c}$, $f_i^{C_2}\in\mathbb{R}^{h_2\times w_2\times c}$ differs, we first align its scale to be consistent resolution (dimension) with $f_i^O\in\mathbb{R}^{h\times w\times c}$ through downsampling. Then we carry out post-processing to $f_i^V, f_i^D, f_i^O, f_i^{C_1}, f_i^{C_2}$ and we only need to post-process the features from different angles ${f_i^V, f_i^D, f_i^O}$ while keeping the resolution unchanged. After post-processing, we cross-concatenate encoded multi-view feature tensors $f_i^V, f_i^D, f_i^O, f_i^{C_1}, f_i^{C_2}$ from same level $i$ (the \textit{mv-tensor} in Fig.~\ref{fig:overview}) to compose one multi-view enhanced feature tensor (the \textit{en-tensor} in Fig.~\ref{fig:overview}). We design a two-stage attention module to enhance feature interaction and correlational clues mining from different views. 

The attention of the first stage aims to aggregate the correlative clues of viewing distance and viewing angle respectively. Taking the feature tensor $\{f_i^D, f_i^V, f_i^O\}$ from three viewing angles as an example, we first compress channel feature through the convolution layer to obtain $f_i^{Ang}\in\mathbb{R}^{h\times w\times c}$, and then input $f_i^{Ang}$ into three parallel tensor multiple modulus multiplication modules to calculate attention. The process is formulated as
\begin{equation}
\begin{aligned}
&f_i^{Ang}=\texttt{ReLU}(\texttt{Conv}(\texttt{Cat}(f_i^D, f_i^V, f_i^O))) \\
&u_A=\sigma(f_i^{Ang}\times_1U_{A_1}\times_2U_{A_2}\times_3U_{A_3}) \\
&u_B=\sigma(f_i^{Ang}\times_1U_{B_1}\times_2U_{B_2}\times_3U_{B_3}) \\
&u_C=\sigma(f_i^{Ang}\times_1U_{C_1}\times_2U_{C_2}\times_3U_{C_3}) \\
&F_i^{Ang}=f_i^D\odot u_A+f_i^V\odot u_B+f_i^O\odot u_C 
\end{aligned}
\end{equation}
where \{$u_A,u_B,u_C$\} are attention factors, $\{f_i^D, f_i^V, f_i^O\}$ indicate feature tensors from three different viewing angles, $F_i^{Ang}$ represents \textit{en-tensor} of the first stage attention, $\sigma$ denotes sigmoid function scaling the weight value into (0, 1). $\texttt{Cat}()$ is the concatenation operation along channel and $\texttt{ReLU}()$ represents the activation function. $\{U_{A_i},U_{B_i},U_{C_i}\}$ represent the parameter matrix of attention factor calculation modules based on tensor multiple modulus multiplication operation, $\times_i$ represents modular multiplication~\cite{TensorDecom}, $\odot$ means element-by-element multiplication. Similarly, we can process the feature tensor $F_i^{Dist}$ of distance-based views after fusion by the same operation. Through such two parallel internal-attention (\textit{In-att}) feature fusion modules, we can enhance the semantic information of the feature maps from different angles and distance.

In the second stage of the attention mechanism, we further interact $F_i^{Ang}$ and $F_i^{Dist}$. As shown in Fig.~\ref{fig:camv}, the features of discriminative viewing angles $F_i^{'Ang}$ is obtained by boundary separation based on self-attention, and $F_i^{'Ang}$ will be used as a complement to $F_i^{Dist}$. Furthermore, we concatenate $F_i^{'Ang}$ and $F_i^{Dist}$ together to obtain the multi-view intermediate feature tensor $F_{MV}$. Finally, we fuse $F_{MV}$ to obtain the final output of CAMV module. The specific process is defined by the following formula:
\begin{equation}
\begin{aligned}
&F_i^{A1}=\texttt{Conv}(\texttt{ReLU}(\texttt{Conv}(\texttt{AvgPool}(F_i^{Ang})))) \\
&F_i^{A2}=\texttt{Conv}(\texttt{ReLU}(\texttt{Conv}(\texttt{MaxPool}(F_i^{Ang})))) \\
&F_i^{Ang}=F_i^{Ang}\odot\sigma(F_i^{A1}+F_i^{A2}) \\
&F_i^{'A1}=\text{avg} (F_i^{Ang}) \\
&F_i^{'A2}=\max(F_i^{Ang}) \\
&F_i^{'Ang}=F_i^{Ang}\odot\sigma(\texttt{Conv}(\texttt{Cat}(F_i^{'A1},F_i^{'A2})))
\end{aligned}
\end{equation}
where $\texttt{MaxPool}()$ and and $\texttt{AvgPool}()$ mean maximum and average pooling respectively, $mean()$ indicates taking the mean of the elements and $max()$ indicates taking the maximum of the elements along the channel dimension. Generally speaking, $\texttt{AvgPool}()$ can preserve more background information, and $\texttt{MaxPool}()$ can preserve more texture information. Thus, abundant boundary information will help to better capture the blurred differences in shape, color, scale and so on between the object and the background.

Through the two-stage attention blocks in CAMV, we carry out implicit interaction and semantic correlation mining for features from different views. The viewing angle and distance well complement the differences between them. The boosted feature expression makes the camouflaged object more clearly separate from the background surroundings. To sum up, CAMV aggregates feature maps from different views, and integrates the auxiliary boundary information into the main branch that incorporates the distance views. Thus, we will transmit a semantically enhanced and more compact feature map into the next processing unit.

\subsection{Channel Fusion Unit}

The input of CFU module is the integrated feature map $f_i$ from CAMV, which is an embedding obtained by fusing features from different views. CFU splits the feature map $f_i$ from CAMV module into $j$ chunks $\{f_i^1,f_i^2,...,f_i^k,...,f_i^j\}$ along the channel dimension, where $k$ indicates the index of different chunks. All chunks $\{f_i^1,f_i^2,...,f_i^k,...,f_i^j\}$ have a consistent number of channels. CFU executes channel-wise local interaction process (CLIP) between adjacent chunks $f_i^{k-1}$ and $f_i^k$ to connect all channels of $f_i^{k-1}$ and $f_i^k$. The output of CLIP is further interacted with the next chunk $f_i^{k+1}$. In this way, all channels of $f_i$ interact with each other. Then, the outputs of all CLIP will be reassembled into one feature map, which will be used as the input of the overall iteration, giving full consideration to the idea of consistency between the overall and the local. The CLIP is described as follows:
\begin{equation}
\textbf{CLIP}(f_i^{k+1}, f_i^k)=\texttt{Tucker}(\texttt{Cat}(f_i^{k+1},\texttt{Conv}(f_i^k)))
\end{equation}
where $\texttt{Tucker}()$ represents the interaction fusion operation based on tensor multiple modulus multiplication, which can filter out the redundant features by splicing and make its semantic expression more compact. 

The overall progressive iteration (OPI), which aims to explore the potential semantic relevance of context, conducts progressive iterations from the overall. This iterative hybrid strategy helps to obtain a more powerful feature representation. The output $z_i$ of the final CLIP is the input of OPI. We define the initial value of $z_i$ as $z_i^0$. For each OPI, 
\begin{equation}
\begin{aligned}
&z_i^0=\texttt{CBR}(z_i) \\
&z_i^{s+1}=\texttt{CBR}(z_i^s+z_i^0)
\end{aligned}
\end{equation}
where $\texttt{CBR}()$ represents a block unit mainly based on convolution layer, including the combination of multiple convolutional layers, batch normalization, and activation layers. We adopt FPN~\cite{FPN} architecture as the feature extractor, which results in multi-level feature maps of different scales. We adopt a progressive upsampling method to gradually restore the feature maps of different levels to be consistent resolution. Finally, a fusion unit and sigmoid function are used to obtain the predicted results.
\begin{figure}
\begin{center}
\includegraphics[width=6cm]{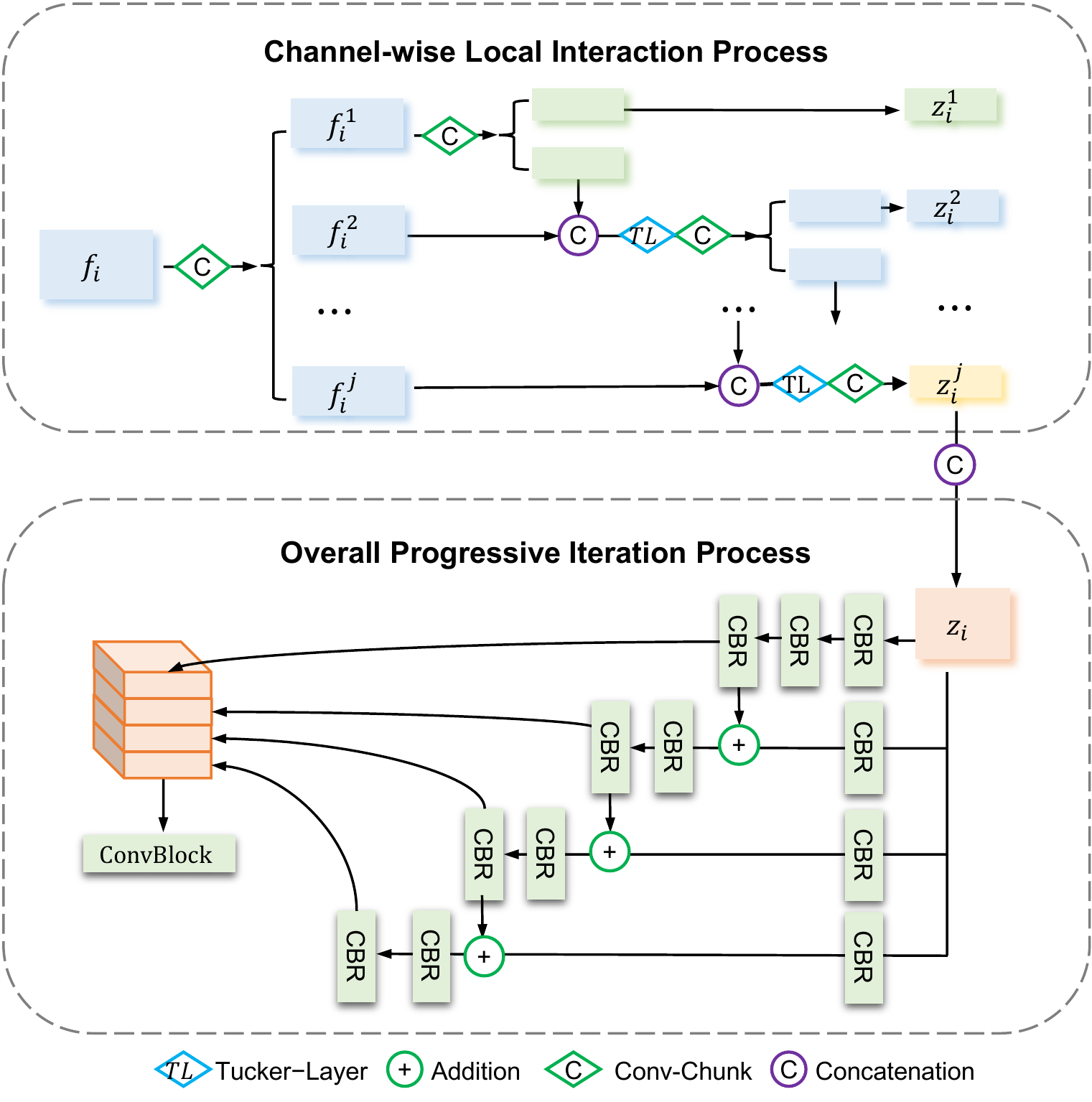}
\end{center}
   \caption{The architecture of the CFU module. CFU first performs feature interaction based on Tucker layer and convolution layer through channel expansion and split, and then obtains the final output through up-down correlation cue mining.}
\label{fig:cfu}
\end{figure}
\begin{table*}[htbp]
  \centering
  \small
  \caption{Comparison of evaluation results of different models on CHAMELEON,COD10K and NC4K. The best model results will be highlighted in \textcolor{dgreen}{\textbf{green}}.}
  \resizebox{\linewidth}{!}{
    \begin{tabular}{c|c|ccccccccccccccc}
    \toprule
    \multirow{2}[2]{*}{Accepted by} & \multirow{2}[2]{*}{Model} & \multicolumn{5}{c}{CHAMELEON}         & \multicolumn{4}{c}{COD10K}    & \multicolumn{6}{c}{NC4K} \\
          &       & $S_m \uparrow$    & $F_\beta^\omega \uparrow$     & $MAE \downarrow$     & $F_\beta \uparrow$     & $E_m \uparrow$     & $S_m \uparrow$     & $F_\beta^\omega \uparrow$     & $MAE \downarrow$     & $F_\beta \uparrow$     & $E_m \uparrow$     & $S_m \uparrow$      & $F_\beta^\omega \uparrow$     & $MAE \downarrow$     & $F_\beta \uparrow$     & $E_m \uparrow$ \\
    \midrule
    \multicolumn{17}{c}{Salient Object Detection / Medical Image Segmentation} \\
    \midrule
    CVPR2018 & PiCANet~\cite{attn_1} & 0.765 & 0.552 & 0.085 & 0.618 & 0.846 & 0.696 & 0.415 & 0.081 & 0.489 & 0.788 & 0.758 & 0.57  & 0.088 & 0.64  & 0.835 \\
    CVPR2019 & BASNet~\cite{edge_1} & 0.847 & 0.771 & 0.044 & 0.795 & 0.894 & 0.661 & 0.432 & 0.071 & 0.486 & 0.749 & 0.695 & 0.546 & 0.095 & 0.61  & 0.785 \\
    CVPR2019 & CPD~\cite{attn_3}   & 0.857 & 0.731 & 0.048 & 0.771 & 0.923 & 0.75  & 0.531 & 0.053 & 0.595 & 0.853 & 0.787 & 0.645 & 0.072 & 0.705 & 0.866 \\
    CVPR2019 & PoolNet~\cite{PoolNet}  & 0.845 & 0.69  & 0.054 & 0.749 & 0.933 & 0.74  & 0.506 & 0.056 & 0.575 & 0.844 & 0.785 & 0.635 & 0.073 & 0.699 & 0.865 \\
    ICCV2019 & EGNet~\cite{edge_2} & 0.797 & 0.649 & 0.065 & 0.702 & 0.884 & 0.736 & 0.517 & 0.061 & 0.582 & 0.854 & 0.777 & 0.639 & 0.075 & 0.696 & 0.864 \\
    AAAI2020 & F3Net~\cite{F3NET} & 0.848 & 0.744 & 0.047 & 0.77  & 0.917 & 0.739 & 0.544 & 0.051 & 0.593 & 0.819 & 0.78  & 0.656 & 0.07  & 0.705 & 0.848 \\
    ICCV2019 & SCRN~\cite{SCRN}  & 0.876 & 0.741 & 0.042 & 0.787 & 0.939 & 0.789 & 0.575 & 0.047 & 0.651 & 0.88  & 0.83  & 0.698 & 0.059 & 0.757 & 0.897 \\
    CVPR2020 & CSNet~\cite{CSNet} & 0.856 & 0.718 & 0.047 & 0.766 & 0.928 & 0.778 & 0.569 & 0.047 & 0.634 & 0.871 & 0.75  & 0.603 & 0.088 & 0.655 & 0.793 \\
    CVPR2020 & SSAL~\cite{SSAL}  & 0.757 & 0.639 & 0.071 & 0.702 & 0.856 & 0.668 & 0.454 & 0.066 & 0.527 & 0.7789 & 0.699 & 0.561 & 0.093 & 0.644 & 0.812 \\
    CVPR2020 & UCNet~\cite{UCNet} & 0.88  & 0.817 & 0.036 & 0.836 & 0.941 & 0.776 & 0.633 & 0.042 & 0.681 & 0.867 & 0.811 & 0.729 & 0.055 & 0.775 & 0.886 \\
    CVPR2020 & MINet~\cite{MINet} & 0.855 & 0.771 & 0.036 & 0.802 & 0.937 & 0.77  & 0.608 & 0.042 & 0.657 & 0.859 & 0.812 & 0.72  & 0.056 & 0.764 & 0.887 \\
    CVPR2020 & ITSD~\cite{ITSD}  & 0.814 & 0.662 & 0.057 & 0.705 & 0.901 & 0.767 & 0.557 & 0.051 & 0.615 & 0.861 & 0.811 & 0.679 & 0.064 & 0.729 & 0.883 \\
    MICCAI2020 & PraNet~\cite{Pranet} & 0.86  & 0.763 & 0.044 & 0.789 & 0.935 & 0.789 & 0.629 & 0.045 & 0.671 & 0.879 & 0.822 & 0.724 & 0.059 & 0.763 & 0.888 \\
    \midrule
    \multicolumn{17}{c}{Camouflaged Object Detection} \\
    \midrule
    CVPR2020 & SINet~\cite{cod} & 0.872 & 0.806 & 0.034 & 0.827 & 0.946 & 0.776 & 0.631 & 0.043 & 0.679 & 0.874 & 0.808 & 0.723 & 0.058 & 0.769 & 0.883 \\
    CVPR2021 & SLSR~\cite{rank_net}  & 0.89  & 0.822 & 0.03  & 0.841 & 0.948 & 0.804 & 0.673 & 0.037 & 0.715 & 0.892 & 0.84  & 0.766 & 0.048 & 0.804 & 0.907 \\
    CVPR2021 & MGL-R~\cite{MGL} & 0.893 & 0.812 & 0.031 & 0.833 & 0.941 & 0.814 & 0.666 & 0.035 & 0.71  & 0.89  & 0.833 & 0.739 & 0.053 & 0.782 & 0.893 \\
    CVPR2021 & PFNet~\cite{context_reasoning} & 0.882 & 0.81  & 0.033 & 0.828 & 0.945 & 0.8   & 0.66  & 0.04  & 0.701 & 0.89  & 0.829 & 0.745 & 0.053 & 0.784 & 0.898 \\
    CVPR2021 & UJSC*~\cite{uncertainty_joint} & 0.891 & 0.833 & .0.030 & 0.847 & 0.955 & 0.809 & 0.684 & 0.035 & 0.721 & 0.891 & 0.842 & 0.771 & 0.047 & 0.806 & 0.907 \\
    IJCAI2021 & C2FNet~\cite{context-cross} & 0.888 & 0.828 & 0.032 & 0.844 & 0.946 & 0.813 & 0.686 & 0.036 & 0.723 & 0.9   & 0.838 & 0.762 & 0.049 & 0.795 & 0.904 \\
    ICCV2021 & UGTR~\cite{uncertainty_transformer} & 0.888 & 0.794 & 0.031 & 0.819 & 0.94  & 0.817 & 0.666 & 0.036 & 0.711 & 0.89  & 0.839 & 0.746 & 0.052 & 0.787 & 0.899 \\
    CVPR2022 & ZoomNet~\cite{ZoomNet-CVPR2022} & 0.902 & 0.845 & 0.023 & 0.864 & 0.958 & 0.838 & 0.729 & 0.029 & 0.766 & 0.911 & 0.853 & 0.784 & 0.043 & 0.818 & 0.912 \\
    \textbf{OURS}  & \textbf{MFFN}  & \textbf{\textcolor{dgreen}{0.905}} & \textbf{\textcolor{dgreen}{0.852}} &  \textbf{\textcolor{dgreen}{0.021}} & \textbf{\textcolor{dgreen}{0.871}} & \textbf{\textcolor{dgreen}{0.963}} & \textbf{\textcolor{dgreen}{0.846}} & \textbf{\textcolor{dgreen}{0.745}} & \textbf{\textcolor{dgreen}{0.028}} & \textbf{\textcolor{dgreen}{0.782}} & \textbf{\textcolor{dgreen}{0.917}} & \textbf{\textcolor{dgreen}{0.856}} & \textbf{\textcolor{dgreen}{0.791}} & \textbf{\textcolor{dgreen}{0.042}} & \textbf{\textcolor{dgreen}{0.827}} & \textbf{\textcolor{dgreen}{0.915}} \\
    \bottomrule
    \end{tabular}%
    }
  \label{tab:compare}%
\end{table*}%
\subsection{Loss Functions}
Binary cross entropy loss~(BECL) is often used in various image segmentation tasks, and its mathematical form is as follows:
\begin{equation}
l_{BCEL}^{i,j}=-g_{i,j}\log p_{i,j}-(1-g_{i,j})\log(1-p_{i,j})
\end{equation}
where $g_{i,j}\in\{0,1\}$ and $p_{i,j}\in[0,1]$ denote the ground truth and the predicted value at position(i,j), respectively. Because the camouflage object is often seriously disturbed by the background surroundings. As a result, the model produces serious fuzziness and uncertainty in prediction. For this reason, we design uncertainty perceived loss (UAL)~\cite{ZoomNet-CVPR2022} as an auxiliary of BCEL to improve the prediction ability of the model for camouflaged objects. And its mathematical form is as follows:
\begin{equation}
l_{UAL}^{i,j}=1-|2p_{i,j}-1|^2
\end{equation}
finally,the total loss function can be written as:
\begin{equation}
L=L_{DCEL}+\lambda L_{UAL}
\end{equation}
We use the UAL form of the quadratic power because the quadratic curve has a gentle gradient around 0 and 1 while maintaining a reasonable penalty interval around 0.5. The cosine strategy is used to dynamically adjust the $\lambda$.

\section{Experiments}
\subsection{Experiment Setup}
\paragraph{Datasets.} We use four COD datasets, CAMO~\cite{CAMO}, CHAMELEON~\cite{chameleon}, COD10K~\cite{cod} and NC4K~\cite{rank_net}. CAMO consists of 1,250 camouflaged and 1,250 non-camouflaged images. CHAMELEON contains 76 hand-annotated images. COD10K includes 5,066 camouflaged, 3,000 background. NC4K is another COD testing dataset including 4,121 images. In this work, we use CAMO and COD10K to construct a training set containing 4,040 camouflage images. To fully verify the generalization ability of the model, we conducted tests on CHAMELEON and NC4K that did not participate in the training, as well as the rest of COD10K. The train, validation, and test sets have been split by default in their original corresponding datasets. 

\paragraph{Evaluation Metrics.} To facilitate comparison with previous methods, we adopt the following evaluation indicators: Structure-measure~($S_m$) which is used as an assessment of structural similarity, F-measure~($F_\beta$) which is used to balance estimates of accuracy and recall rates, weighted F-measure~($F_\beta^w$), mean absolute error~(MAE), Enhanced-alignment measure~($E_m$), which considers both the global average of image and local pixel matching.

\paragraph{Experiment environment.} The entire mod code is implemented based on PyTorch. In the feature extraction stage, ResNet-50 model pretrained on ImageNet is adopted.

\paragraph{Hyperparameter setting.} To verify the validity of the model itself, we followed the same hyperparameter settings as most of the comparison models. SGD with a momentum of 0.9 and a weight decay of 0.0005 was chosen as the optimizer. We initialize the learning rate to 0.01 and follow the cosine preheat decay strategy. In addition, we set batch size to 8, we trained our model in the training set, and evaluated it in the independent validation set every three epochs. When 60\% results of the evaluation metrics of the model on the validation set did not exceed the previous evaluation results, the training was stopped. For more detailed information, please 
see Appendix D.

\subsection{Comparisons with State-of-the-Arts}
Due to the wide application value of COD, researchers have published a number of models with excellent performance in top academic conferences and journals. We selected models that have appeared in mainstream computer vision conferences in recent years for comparison and applied the published results. In addition, during the training process, we follow the same backbone and hyperparameters' settings (\ie batch size, learning rate, and optimizer) as most models.

\begin{figure}
    \centering
    \begin{subfigure}[b]{0.4\textwidth}
         \centering
         \includegraphics[width=0.9\textwidth]{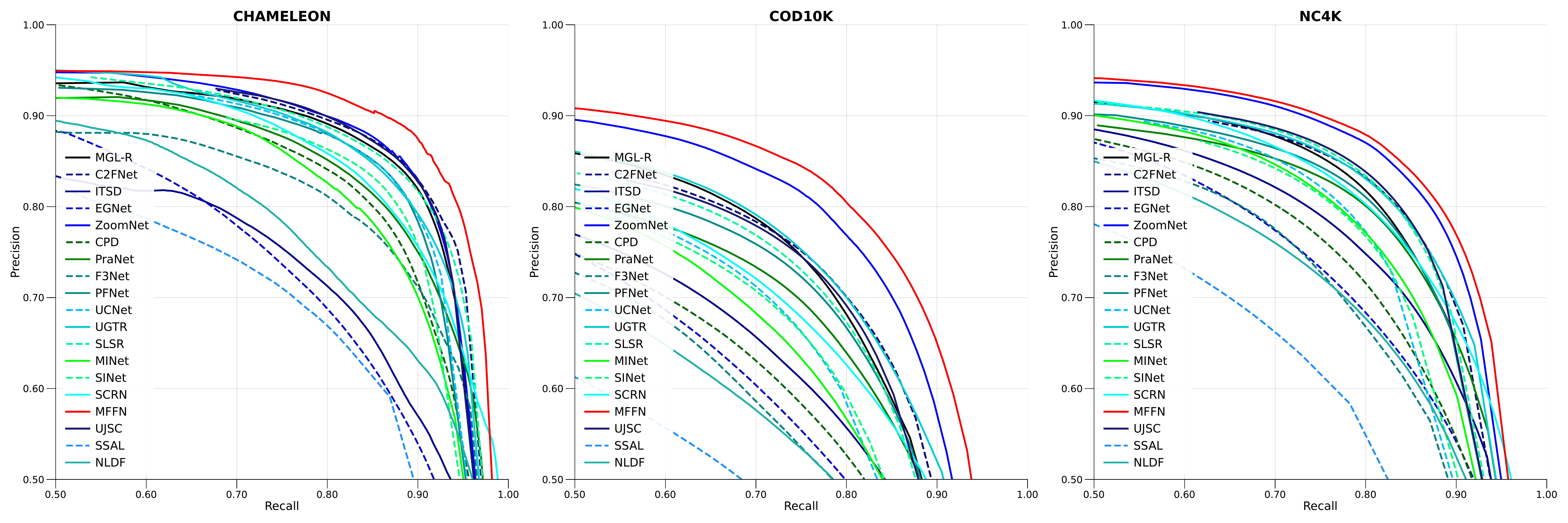}
         \caption{PR curves of the proposed MFFN and recent SOTA algorithms over CHAMELEON, COD10K and NC4K.}
         \label{PR curves}
     \end{subfigure}
     \hfill
    \begin{subfigure}[b]{0.4\textwidth}
         \centering
         \includegraphics[width=0.9\textwidth]{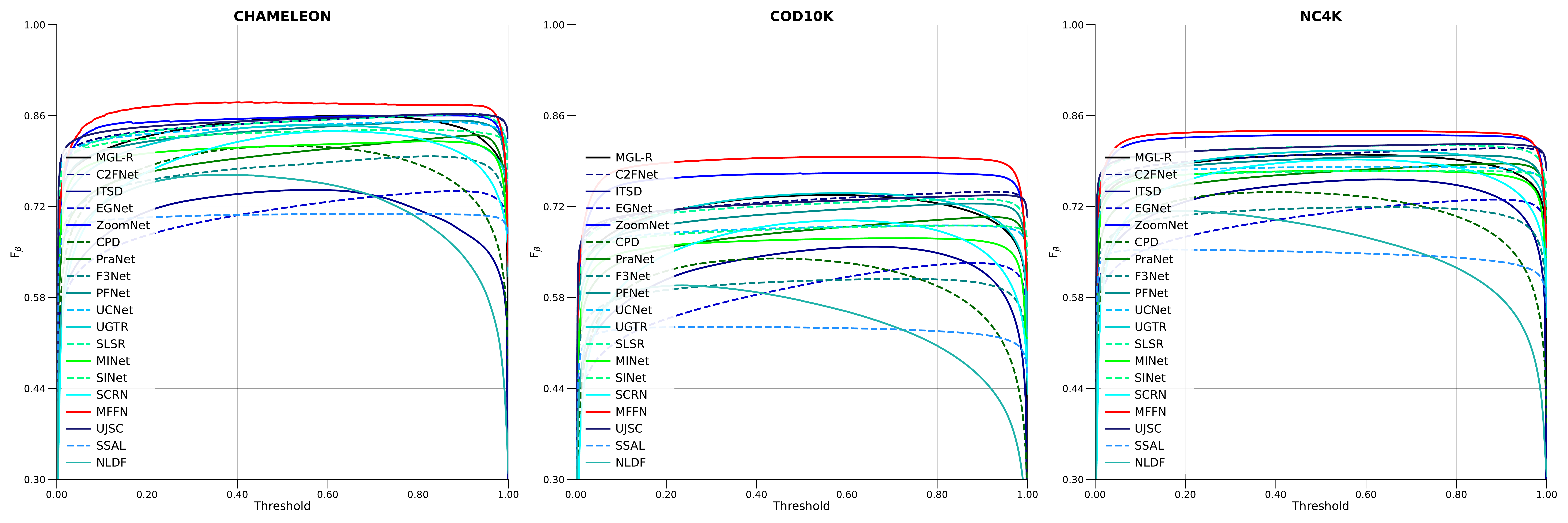}
         \caption{$F_\beta$ curves of the proposed MFFN and recent SOTA algorithms over CHAMELEON, COD10K and NC4K.}
         \label{beta_curves}
     \end{subfigure}
     \caption{Results of PR and $F_\beta$ curves. \textcolor{red}{Red line} represents our proposed MFFN.}
\end{figure}

\begin{figure}
\begin{center}
\includegraphics[width=8cm]{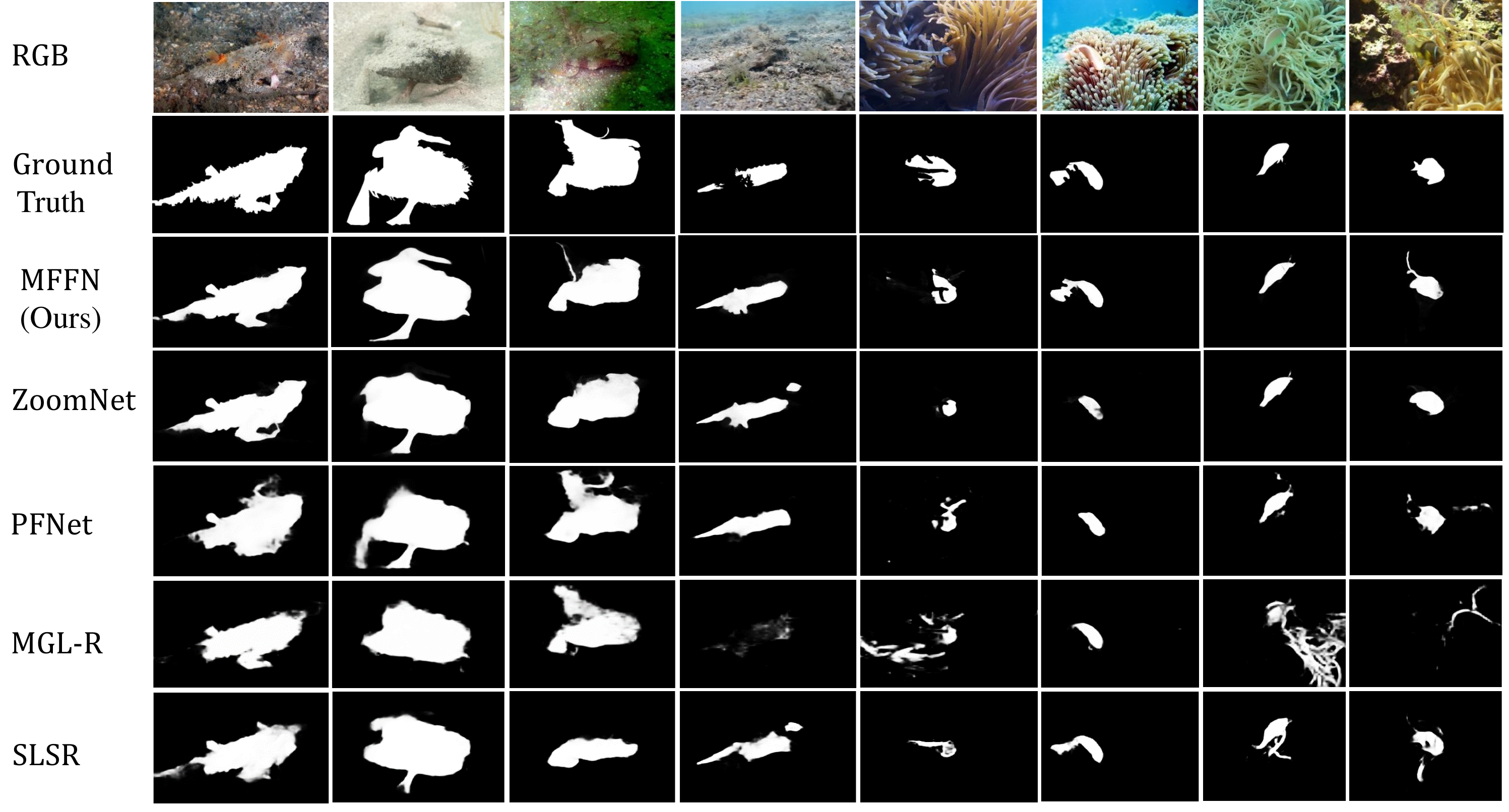}
\end{center}
   \caption{Visual comparisons of some latest research algorithms and our proposed MFFN in some typical images. We can find that the prediction results of MFFN have clearer boundary and region shape.}
\label{fig:visual}
\end{figure}

As shown in Tab.~\ref{tab:compare}, MFFN achieves the best performance in all three test datasets without extra training data, especially for the four indicators of $S_m$, $F_\beta$, $F_\beta^w$, $E_m$. Compared with model MGL~\cite{MGL}, which introduced boundary detection as an auxiliary task with interactive learning and graph neural network, it is obvious that our method has superior performance by capturing boundary information and separating background simultaneously. Compared with ZoomNet~\cite{ZoomNet-CVPR2022} which achieved the second best performance, our model improves $S_m$ by 0.8\%, $F_\beta^w$ by 1.6\%, $F_\beta$ by 1.6\% and $E_m$ by 0.6\% in the COD10K test set. Similarly, in the NC4K dataset test results, $S_m$ is improved by 0.3\%, $F_\beta^w$ is improved by 0.7\%, $F_\beta$ is improved by 0.9\% and $E_m$ is improved by 0.3\%. We draw precision-recall (PR) curve and $F_\beta$ curve. As shown in Fig.~\ref{PR curves} and Fig.~\ref{beta_curves}, the PR curve of MFFN surrounds the previously proposed model, and the $F_\beta$ curve also presents an almost horizontal shape, which represents that MFFN has more accurate detection results. The visualization results for the different methods are shown in Fig.~\ref{fig:visual}. We select 8 samples with obvious differences in object size, background interference and color for analysis. The comparison results show that our method can still obtain clear prediction boundaries and region shapes under the circumstance of highly blurred boundary and highly similar background. For model complexity, although we increase the input images with the multi-view design, our model still has the least number of parameters compared with single-view models, as shown in Tab. \ref{tab:complex}. This indicates that with multi-view design, we are able to apply a simpler encoder (\ie instead of ViT~\cite{uncertainty_transformer}) with less complex strategies (\ie instead of joint SOD and COD~\cite{uncertainty_joint}, or joint mutual graph learning~\cite{MGL}). 

\begin{table}[htbp]
  \centering
  \caption{Comparison of the number of parameters of our proposed MFFN and other SOTA models.}
  
  \resizebox{\linewidth}{!}{
    \begin{tabular}{c|ccccccc}
    \toprule
    Method  & MFFN(Ours)  & UGTR~\cite{uncertainty_transformer}    & UJSC~\cite{uncertainty_joint}    & ZoomNet~\cite{ZoomNet-CVPR2022}     & PfNet    ~\cite{context_reasoning}& MGL-R~\cite{MGL} & SLSR~\cite{rank_net} \\
    \midrule
    Parameters   & 36.554M & 48.868M & 217.982M & 32.382M & 46.498M & 63.595M & 50.935M \\
    \bottomrule
    \end{tabular}%
  \label{tab:complex}%
  }
\end{table}%


\subsection{Ablation Studies}

In this section, we conduct ablation studies on the combination of different views, the mode to interact of multiple views and the CFU. Considering the representativeness of the dataset, different model design methods were used to evaluate the two large datasets COD10K and NC4K.

\paragraph{The effect of different views.} The proposed model aims to complement and fuse the multi-view information to obtain precise and accurate boundary information and semantic correlation. We expand on the \textit{distance view}, \textit{perspective view}, \textit{angle view} and the different combinations of them. The experimental results shown in the Tab.~\ref{tab:cod10ks} and Tab.~\ref{tab:nc4ks} reveal the significance of different views for feature capture, and we choose the best combination of views.

\paragraph{The effect of two-stage attention in CAMV.} In our method, we introduce CAMV to interact with multi-view feature maps, enhancing the semantic expression of foreground and background. In order to better analyze the effect of two-stage attention on model performance improvement, we analyze the two stages respectively.

\begin{table}[htbp]
  \centering
  \caption{Comparisons of different views and their combinations using different CAMV on COD10K. V-O: original view; V-F: far view; V-C: close view; V-A: angle view; V-P: perspective view.}
  \resizebox{0.8\linewidth}{!}{
    \begin{tabular}{c|c|ccccc}
    \toprule
    View  & CAMV  & $S_m \uparrow$    & $F_\beta^\omega \uparrow$     & $MAE \downarrow$     & $F_\beta \uparrow$     & $E_m \uparrow$ \\
    \midrule
    V-O   &       -- & 0.797 & 0.649 & 0.063 & 0.704 & 0.875 \\
    V-F   & One-stage & 0.808 & 0.678 & 0.033 & 0.721 & 0.884 \\
    V-C   & One-stage & 0.841 & 0.736 & 0.029 & 0.772 & 0.909 \\
    V-A   & One-stage & 0.812 & 0.686 & 0.034 & 0.729 & 0.892 \\
    \multirow{2}[0]{*}{V-F\&C} & One-stage & 0.844 & 0.735 & 0.028 & 0.769 & 0.907 \\
          & Two-stage & 0.842 & 0.735 & 0.028 & 0.771 & 0.911 \\
    \multirow{2}[0]{*}{V-A\&F} & One-stage & 0.807 & 0.675 & 0.034 & 0.717 & 0.886 \\
          & Two-stage & 0.805 & 0.673 & 0.035 & 0.717 & 0.882 \\
    \multirow{2}[1]{*}{V-C\&P} & One-stage & 0.827 & 0.714 & 0.036  & 0.759 & 0.901 \\
          & Two-stage & 0.838 &	0.725&	0.031&	0.764&	0.907 \\
    \multirow{2}[1]{*}{V-A\&P} & One-stage & 0.796 & 0.649 & 0.042  & 0.689 & 0.881 \\
          & Two-stage & 0.802&	0.660&	0.037&	0.707&	0.886 \\
    \multirow{2}[1]{*}{V-A\&C} & One-stage & 0.835 & 0.727 & 0.03  & 0.766 & 0.906 \\
          & Two-stage & 0.846 & 0.745 & 0.028 & 0.782 & 0.917 \\
    \bottomrule
    \end{tabular}%
  \label{tab:cod10ks}%
  }
\end{table}%
\begin{table}[htbp]
  \centering
  \caption{Influence of CFU module on performance.}
  \resizebox{0.8\linewidth}{!}{
    \begin{tabular}{c|c|ccccc}
    \toprule
    Dataset & Method & $S_m \uparrow$    & $F_\beta^\omega \uparrow$     & $MAE \downarrow$     & $F_\beta \uparrow$     & $E_m \uparrow$ \\
    \midrule
    \multirow{2}[1]{*}{COD10K} & no-CFU & 0.844 & 0.73  & 0.03  & 0.771 & 0.917 \\
          & CFU   & 0.846 & 0.745 & 0.028 & 0.782 & 0.917 \\
    \multirow{2}[1]{*}{NC4K} & no-CFU & 0.854 & 0.78  & 0.045 & 0.819 & 0.915 \\
          & CFU   & 0.856 & 0.791 & 0.042 & 0.827 & 0.915 \\
    \bottomrule
    \end{tabular}%
    }
  \label{tab:cfure}%
\end{table}%
\begin{table}[htbp]
  \centering
  \caption{Comparisons of different views and their combinations using different CAMV on NC4K.}
  \resizebox{0.8\linewidth}{!}{
    \begin{tabular}{c|c|ccccc}
    \toprule
    View  & CAMV  & $S_m \uparrow$    & $F_\beta^\omega \uparrow$     & $MAE \downarrow$     & $F_\beta \uparrow$     & $E_m \uparrow$ \\
    \midrule
    V-F   & One-stage & 0.839 & 0.764 & 0.046 & 0.801 & 0.902 \\
    V-C   & One-stage & 0.854 & 0.788 & 0.043 & 0.823 & 0.912 \\
    V-A   & One-stage & 0.839 & 0.764 & 0.047 & 0.802 & 0.903 \\
    \multirow{2}[0]{*}{V-F\&C} & One-stage & 0.86  & 0.793 & 0.042 & 0.824 & 0.914 \\
          & Two-stage & 0.857 & 0.79  & 0.042 & 0.823 & 0.913 \\
    \multirow{2}[0]{*}{V-A\&F} & One-stage & 0.834 & 0.757 & 0.048 & 0.795 & 0.9 \\
          & Two-stage & 0.833 & 0.755 & 0.049 & 0.795 & 0.9 \\
    \multirow{2}[1]{*}{V-C\&P} & One-stage & 0.843 & 0.774 & 0.049 & 0.806 & 0.897 \\
          & Two-stage & 0.852&	0.782&	0.046&	0.817&	0.909 \\
    \multirow{2}[1]{*}{V-A\&P} & One-stage & 0.821 & 0.742 & 0.054 & 0.780 & 0.886 \\
          & Two-stage & 0.835&	0.753&	0.050&	0.792&	0.899 \\
    \multirow{2}[1]{*}{V-A\&C} & One-stage & 0.845 & 0.774 & 0.047 & 0.812 & 0.906 \\
          & Two-stage & 0.856 & 0.791 & 0.042 & 0.827 & 0.915 \\
    \bottomrule
    \end{tabular}%
    }
  \label{tab:nc4ks}%
\end{table}%
\paragraph{The effect of CFU.} Considering the deficiency in context semantic association of feature maps after multi-view fusion, we design the CFU to further potential mine clues of the feature tensors obtained after CAMV. CFU module mainly includes channel expansion and interaction and context extraction modules. We perform an ablation analysis on the two main parts of the two CFU. Based on the results in the Tab.~\ref{tab:cfure}, it is discovered that obtaining potential context clues through CFU is critical.

\section{Conclusion}
In this paper, we propose the MFFN model by imitating the multi-view observation mechanism of biology, which makes the features captured from different views complement and interact with each other. MFFN makes up for the omission of features in fixed view observation. Firstly, we obtain more compact features through multi-view attentional interaction design, which enhances the semantic representation ability of the feature maps to the object region and boundary, and well integrates the multi-view semantic information. In addition, the context association information of feature tensor, which is implied in the channel dimension, is further mined by the CFU. A large number of experimental results verify the high performance of this method in COD task, which is superior to the previous method. MFFN shows SOTA results in the COD task and is equally good in the SOD task, but our multi-view design concept still needs further development to achieve accurate detection performance in general object detection tasks.

{\small
\bibliographystyle{ieee_fullname}
\bibliography{wacv_0172}

\begin{thebibliography}{10}\itemsep=-1pt

\bibitem{Fb}
Radhakrishna Achanta, Sheila Hemami, Francisco Estrada, and Sabine Susstrunk.
\newblock Frequency-tuned salient region detection.
\newblock In {\em Proceedings of the IEEE Conference on Computer Vision and
  Pattern Recognition (CVPR)}, June 2009.

\bibitem{color_1}
Ali Borji.
\newblock What is a salient object? a dataset and a baseline model for salient
  object detection.
\newblock {\em IEEE Transactions on Image Processing}, 24(2):742--756, 2015.

\bibitem{gcpanet}
Zuyao Chen, Qianqian Xu, Runmin Cong, and Qingming Huang.
\newblock Global context-aware progressive aggregation network for salient
  object detection.
\newblock {\em Proceedings of the AAAI Conference on Artificial Intelligence},
  34(07):10599--10606, Apr. 2020.

\bibitem{Camouflage_mechanisms}
R.~C. Duarte, Aav Flores, and M. Stevens.
\newblock Camouflage through colour change: mechanisms, adaptive value and
  ecological significance.
\newblock {\em Philosophical Transactions of the Royal Society of London},
  372(1724):75–92, 2017.

\bibitem{Pranet}
Dengping Fan, Gepeng Ji, Tao Zhou, Geng Chen, Huazhong Fu, Shen Jianbing, and
  Ling Shao.
\newblock Pranet: Parallel reverse attention network for polyp segmentation.
\newblock In {\em International Conference on Medical Image Computing and
  Computer-Assisted Intervention (MICCAI)}, October 2020.

\bibitem{cod}
Dengping Fan, Ge-Peng Ji, Guolei Sun, Ming-Ming Cheng, Jianbing Shen, and Ling
  Shao.
\newblock Camouflaged object detection.
\newblock In {\em Proceedings of the IEEE/CVF Conference on Computer Vision and
  Pattern Recognition (CVPR)}, June 2020.

\bibitem{Sm}
Deng-Ping Fan, Ming-Ming Cheng, Yun Liu, Tao Li, and Ali Borji.
\newblock Structure-measure: A new way to evaluate foreground maps.
\newblock In {\em Proceedings of the IEEE International Conference on Computer
  Vision (ICCV)}, October 2017.

\bibitem{Em}
Deng-Ping Fan, Cheng Gong, Yang Cao, Bo Ren, Ming-Ming Cheng, and Ali Borji.
\newblock Enhanced-alignment measure for binary foreground map evaluation.
\newblock In {\em Proceedings of the 27th International Joint Conference on
  Artificial Intelligence (IJCAI)}, July 2018.

\bibitem{attn_5}
Jun Fu, Jing Liu, Haijie Tian, Yong Li, Yongjun Bao, Zhiwei Fang, and Hanqing
  Lu.
\newblock Dual attention network for scene segmentation.
\newblock In {\em Proceedings of the IEEE Conference on Computer Vision and
  Pattern Recognition (CVPR)}, June 2019.

\bibitem{RDL}
Fuentes, Yoon, Kim, SC, Park, and DS.
\newblock A robust deep-learning-based detector for real-time tomato plant
  diseases and pests recognition.
\newblock {\em SENSORS-BASEL}, 17:582–596, 2019.

\bibitem{CSNet}
Shanghua Gao, Yongqiang Tan, Mingming Cheng, Chengze Lu, Yunpeng Chen, and
  Shuicheng Yan.
\newblock Highly efficient salient object detection with 100k parameters.
\newblock In Andrea Vedaldi, Horst Bischof, Thomas Brox, and Jan-Michael Frahm,
  editors, {\em Proceedings of the European Conference on Computer Vision
  (ECCV)}, August 2020.

\bibitem{resnet}
Kaiming He, Xiangyu Zhang, Shaoqing Ren, and Jian Sun.
\newblock Deep residual learning for image recognition.
\newblock In {\em Proceedings of the IEEE Conference on Computer Vision and
  Pattern Recognition (CVPR)}, June 2016.

\bibitem{DLT}
M.~H. Hesamian, Wenjing Jia, Xiangjian He, and Paul Kennedy.
\newblock Deep learning techniques for medical image segmentation: Achievements
  and challenges.
\newblock {\em Journal of Digital Imaging}, 32:582–596, 2019.

\bibitem{M_scal_3}
Qibin Hou, Ming-Ming Cheng, Xiaowei Hu, Ali Borji, Zhuowen Tu, and Philip H.~S.
  Torr.
\newblock Deeply supervised salient object detection with short connections.
\newblock {\em IEEE Transactions on Pattern Analysis and Machine Intelligence},
  41(4):815--828, 2019.

\bibitem{ji2022deep}
Ge-Peng Ji, Deng-Ping Fan, Yu-Cheng Chou, Dengxin Dai, Alexander Liniger, and
  Luc Van~Gool.
\newblock Deep gradient learning for efficient camouflaged object detection.
\newblock {\em arXiv preprint arXiv:2205.12853}, 2022.

\bibitem{contrast_2}
Huaizu Jiang, Jingdong Wang, Zejian Yuan, Yang Wu, Nanning Zheng, and Shipeng
  Li.
\newblock Salient object detection: A discriminative regional feature
  integration approach.
\newblock In {\em Proceedings of the IEEE Conference on Computer Vision and
  Pattern Recognition (CVPR)}, June 2013.

\bibitem{TensorDecom}
T. Kolda and B. Bader.
\newblock Tensor decompositions and applications.
\newblock {\em Siam Review}, 51:455--500, 2009.

\bibitem{CAMO}
Trung-Nghia Le, Tam~V. Nguyen, Zhongliang Nie, Minh-Triet Tran, and Akihiro
  Sugimoto.
\newblock Anabranch network for camouflaged object segmentation.
\newblock {\em Computer Vision and Image Understanding}, 184:45--56, 2019.

\bibitem{M_scal_5}
Gayoung Lee, Yu-Wing Tai, and Junmo Kim.
\newblock Deep saliency with encoded low level distance map and high level
  features.
\newblock In {\em Proceedings of the IEEE Conference on Computer Vision and
  Pattern Recognition (CVPR)}, June 2016.

\bibitem{uncertainty_joint}
Aixuan Li, Jing Zhang, Yunqiu Lv, Bowen Liu, Tong Zhang, and Yuchao Dai.
\newblock Uncertainty-aware joint salient object and camouflaged object
  detection.
\newblock In {\em Proceedings of the IEEE/CVF Conference on Computer Vision and
  Pattern Recognition (CVPR)}, June 2021.

\bibitem{HKUIS}
Guanbin Li and Yizhou Yu.
\newblock Visual saliency based on multiscale deep features.
\newblock In {\em Proceedings of the IEEE Conference on Computer Vision and
  Pattern Recognition (CVPR)}, June 2015.

\bibitem{MDF}
Guanbin Li and Yizhou Yu.
\newblock Visual saliency detection based on multiscale deep cnn features.
\newblock {\em IEEE Transactions on Image Processing}, 25(11):5012--5024, 2016.

\bibitem{color_2}
Yin Li, Xiaodi Hou, Christof Koch, James~M. Rehg, and Alan~L. Yuille.
\newblock The secrets of salient object segmentation.
\newblock In {\em Proceedings of the IEEE Conference on Computer Vision and
  Pattern Recognition (CVPR)}, June 2014.

\bibitem{FPN}
Tsung-Yi Lin, Piotr Dollar, Ross Girshick, Kaiming He, Bharath Hariharan, and
  Serge Belongie.
\newblock Feature pyramid networks for object detection.
\newblock In {\em Proceedings of the IEEE Conference on Computer Vision and
  Pattern Recognition (CVPR)}, July 2017.

\bibitem{M_scal_1}
Xiao Lin, Zhi-Jie Wang, Lizhuang Ma, and Xiabao Wu.
\newblock Saliency detection via multi-scale global cues.
\newblock {\em IEEE Transactions on Multimedia}, 21(7):1646--1659, 2019.

\bibitem{aleatoric_uncertainty}
Jiawei Liu, Jing Zhang, and Nick Barnes.
\newblock Modeling aleatoric uncertainty for camouflaged object detection.
\newblock In {\em Proceedings of the IEEE/CVF Winter Conference on Applications
  of Computer Vision (WACV)}, January 2022.

\bibitem{PoolNet}
Jiang-Jiang Liu, Qibin Hou, Ming-Ming Cheng, Jiashi Feng, and Jianmin Jiang.
\newblock A simple pooling-based design for real-time salient object detection.
\newblock In {\em Proceedings of the IEEE Conference on Computer Vision and
  Pattern Recognition (CVPR)}, June 2019.

\bibitem{attn_1}
Nian Liu, Junwei Han, and Ming-Hsuan Yang.
\newblock Picanet: Learning pixel-wise contextual attention for saliency
  detection.
\newblock In {\em Proceedings of the IEEE Conference on Computer Vision and
  Pattern Recognition (CVPR)}, June 2018.

\bibitem{picanet}
Nian Liu, Junwei Han, and Ming-Hsuan Yang.
\newblock Picanet: Learning pixel-wise contextual attention for saliency
  detection.
\newblock In {\em Proceedings of the IEEE Conference on Computer Vision and
  Pattern Recognition (CVPR)}, June 2018.

\bibitem{vst}
Nian Liu, Ni Zhang, Kaiyuan Wan, Ling Shao, and Junwei Han.
\newblock Visual saliency transformer.
\newblock In {\em Proceedings of the IEEE/CVF International Conference on
  Computer Vision (ICCV)}, pages 4722--4732, October 2021.

\bibitem{samnet}
Yun Liu, Xin-Yu Zhang, Jia-Wang Bian, Le Zhang, and Ming-Ming Cheng.
\newblock Samnet: Stereoscopically attentive multi-scale network for
  lightweight salient object detection.
\newblock {\em IEEE Transactions on Image Processing}, 30:3804--3814, 2021.

\bibitem{rank_net}
Yunqiu Lv, Jing Zhang, Yuchao Dai, Aixuan Li, Bowen Liu, Nick Barnes, and
  Deng-Ping Fan.
\newblock Simultaneously localize, segment and rank the camouflaged objects.
\newblock In {\em Proceedings of the IEEE/CVF Conference on Computer Vision and
  Pattern Recognition (CVPR)}, June 2021.

\bibitem{Fbw}
Ran Margolin, Lihi Zelnik-Manor, and Ayellet Tal.
\newblock How to evaluate foreground maps.
\newblock In {\em Proceedings of the IEEE Conference on Computer Vision and
  Pattern Recognition (CVPR)}, June 2014.

\bibitem{context_reasoning}
Haiyang Mei, Ge-Peng Ji, Ziqi Wei, Xin Yang, Xiaopeng Wei, and Deng-Ping Fan.
\newblock Camouflaged object segmentation with distraction mining.
\newblock In {\em Proceedings of the IEEE/CVF Conference on Computer Vision and
  Pattern Recognition (CVPR)}, pages 8772--8781, June 2021.

\bibitem{ZoomNet-CVPR2022}
Youwei Pang, Xiaoqi Zhao, Tian-Zhu Xiang, Lihe Zhang, and Huchuan Lu.
\newblock Zoom in and out: A mixed-scale triplet network for camouflaged object
  detection.
\newblock In {\em Proceedings of the IEEE/CVF Conference on Computer Vision and
  Pattern Recognition (CVPR)}, June 2022.

\bibitem{MINet}
Youwei Pang, Xiaoqi Zhao, Lihe Zhang, and Huchuan Lu.
\newblock Multi-scale interactive network for salient object detection.
\newblock In {\em Proceedings of the IEEE Conference on Computer Vision and
  Pattern Recognition (CVPR)}, June 2020.

\bibitem{attn_4}
Jongchan Park, Sanghyun Woo, Joon-Young Lee, and In~So Kweon.
\newblock Bam: Bottleneck attention module.
\newblock In {\em British Machine Vision Conference (BMVC)}, September 2018.

\bibitem{pei2022osformer}
Jialun Pei, Tianyang Cheng, Deng-Ping Fan, He Tang, Chuanbo Chen, and Luc
  Van~Gool.
\newblock Osformer: One-stage camouflaged instance segmentation with
  transformers.
\newblock In {\em Proceedings of the European Conference on Computer Vision
  (ECCV)}. Springer, 2022.

\bibitem{contrast_1}
Federico Perazzi, Philipp Krähenbühl, Yael Pritch, and Alexander Hornung.
\newblock Saliency filters: Contrast based filtering for salient region
  detection.
\newblock In {\em Proceedings of the IEEE Conference on Computer Vision and
  Pattern Recognition (CVPR)}, June 2012.

\bibitem{edge_1}
Xuebin Qin, Zichen Zhang, Chenyang Huang, Chao Gao, Masood Dehghan, and Martin
  Jagersand.
\newblock Basnet: Boundary-aware salient object detection.
\newblock In {\em 2019 IEEE/CVF Conference on Computer Vision and Pattern
  Recognition (CVPR)}, June 2019.

\bibitem{UNet3}
Olaf Ronneberger, Philipp Fischer, and Thomas Brox.
\newblock U-net: Convolutional networks for biomedical image segmentation.
\newblock In {\em International Conference on Medical Image Computing and
  Computer-Assisted Intervention (MICCAI)}, October 2015.

\bibitem{chameleon}
P Skurowski, H Abdulameer, J Baszczyk, T Depta, A Kornacki, and P Kozie.
\newblock Animal camouflage analysis: Chameleon database.
\newblock In {\em Unpublished Manuscript}, 2018.

\bibitem{Camouflage}
Martin Stevens and Sami Merilaita.
\newblock Animal camouflage: current issues and new perspectives.
\newblock {\em Philosophical Transactions of the Royal Society B: Biological
  Sciences}, 364(1516):423--427, 2009.

\bibitem{edge_3}
Jinming Su, Jia Li, Yu Zhang, Changqun Xia, and Yonghong Tian.
\newblock Selectivity or invariance: Boundary-aware salient object detection.
\newblock In {\em Proceedings of the IEEE Conference on Computer Vision and
  Pattern Recognition (CVPR)}, June 2019.

\bibitem{context-cross}
Yujia Sun, Geng Chen, Tao Zhou, Yi Zhang, and Nian Liu.
\newblock Context-aware cross-level fusion network for camouflaged object
  detection.
\newblock In {\em Proceedings of the Thirtieth International Joint Conference
  on Artificial Intelligence, {IJCAI-21}}, August 2021.

\bibitem{cvp}
Tom Troscianko, Benton~Christopher P, Lovell~P. George, Tolhurst~David J, and
  Pizlo Zygmunt.
\newblock Camouflage and visual perception.
\newblock {\em Philosophical Transactions of the Royal Society B: Biological
  Sciences}, 2009.

\bibitem{duts}
Lijun Wang, Huchuan Lu, Yifan Wang, Mengyang Feng, Dong Wang, Baocai Yin, and
  Xiang Ruan.
\newblock Learning to detect salient objects with image-level supervision.
\newblock In {\em Proceedings of the IEEE Conference on Computer Vision and
  Pattern Recognition (CVPR)}, July 2017.

\bibitem{attn_6}
Qilong Wang, Banggu Wu, Pengfei Zhu, Peihua Li, Wangmeng Zuo, and Qinghua Hu.
\newblock Eca-net: Efficient channel attention for deep convolutional neural
  networks.
\newblock In {\em Proceedings of the IEEE Conference on Computer Vision and
  Pattern Recognition (CVPR)}, June 2020.

\bibitem{pagenet}
Wenguan Wang, Shuyang Zhao, Jianbing Shen, Steven C.~H. Hoi, and Ali Borji.
\newblock Salient object detection with pyramid attention and salient edges.
\newblock In {\em Proceedings of the IEEE/CVF Conference on Computer Vision and
  Pattern Recognition (CVPR)}, June 2019.

\bibitem{F3NET}
Jun Wei, Shuhui Wang, and Qingming Huang.
\newblock F3net: Fusion, feedback and focus for salient object detection.
\newblock {\em AAAI Conference on Artificial Intelligence (AAAI)}, February
  2020.

\bibitem{attn_3}
Zhe Wu, Li Su, and Qingming Huang.
\newblock Cascaded partial decoder for fast and accurate salient object
  detection.
\newblock In {\em Proceedings of the IEEE Conference on Computer Vision and
  Pattern Recognition (CVPR)}, June 2019.

\bibitem{SCRN}
Zhe Wu, Li Su, and Qingming Huang.
\newblock Stacked cross refinement network for edge-aware salient object
  detection.
\newblock In {\em Proceedings of the IEEE Conference on Computer Vision and
  Pattern Recognition (CVPR)}, June 2019.

\bibitem{sgl}
Binwei Xu, Haoran Liang, Ronghua Liang, and Peng Chen.
\newblock Locate globally, segment locally: A progressive architecture with
  knowledge review network for salient object detection.
\newblock {\em Proceedings of the AAAI Conference on Artificial Intelligence},
  35(4):3004--3012, May 2021.

\bibitem{texture_1}
Qiong Yan, Li Xu, Jianping Shi, and Jiaya Jia.
\newblock Hierarchical saliency detection.
\newblock In {\em Proceedings of the IEEE Conference on Computer Vision and
  Pattern Recognition (CVPR)}, June 2013.

\bibitem{ECSSD}
Qiong Yan, Li Xu, Jianping Shi, and Jiaya Jia.
\newblock Hierarchical saliency detection.
\newblock In {\em Proceedings of the IEEE Conference on Computer Vision and
  Pattern Recognition (CVPR)}, June 2013.

\bibitem{uncertainty_transformer}
Fan Yang, Qiang Zhai, Xin Li, Rui Huang, Ao Luo, Hong Cheng, and Deng-Ping Fan.
\newblock Uncertainty-guided transformer reasoning for camouflaged object
  detection.
\newblock In {\em Proceedings of the IEEE/CVF International Conference on
  Computer Vision (ICCV)}, October 2021.

\bibitem{hrs}
Yi Zeng, Pingping Zhang, Jianming Zhang, Zhe Lin, and Huchuan Lu.
\newblock Towards high-resolution salient object detection.
\newblock In {\em Proceedings of the IEEE/CVF International Conference on
  Computer Vision (ICCV)}, October 2019.

\bibitem{MGL}
Qiang Zhai, Xin Li, Fan Yang, Chenglizhao Chen, Hong Cheng, and Deng-Ping Fan.
\newblock Mutual graph learning for camouflaged object detection.
\newblock In {\em Proceedings of the IEEE/CVF Conference on Computer Vision and
  Pattern Recognition (CVPR)}, June 2021.

\bibitem{UCNet}
Jing Zhang, Deng-Ping Fan, Yuchao Dai, Saeed Anwar, Fatemeh~Sadat Saleh, Tong
  Zhang, and Nick Barnes.
\newblock Uc-net: Uncertainty inspired rgb-d saliency detection via conditional
  variational autoencoders.
\newblock In {\em Proceedings of the IEEE Conference on Computer Vision and
  Pattern Recognition (CVPR)}, 2020.

\bibitem{SSAL}
Jing Zhang, Xin Yu, Aixuan Li, Peipei Song, Bowen Liu, and Yuchao Dai.
\newblock Weakly-supervised salient object detection via scribble annotations.
\newblock In {\em Proceedings of the IEEE Conference on Computer Vision and
  Pattern Recognition (CVPR)}, June 2020.

\bibitem{automsfnet}
Miao Zhang, Tingwei Liu, Yongri Piao, Shunyu Yao, and Huchuan Lu.
\newblock Auto-msfnet: Search multi-scale fusion network for salient object
  detection.
\newblock In {\em Proceedings of the 29th ACM International Conference on
  Multimedia}, MM '21, page 667–676. Association for Computing Machinery,
  2021.

\bibitem{attn_2}
Xiaoning Zhang, Tiantian Wang, Jinqing Qi, Huchuan Lu, and Gang Wang.
\newblock Progressive attention guided recurrent network for salient object
  detection.
\newblock In {\em Proceedings of the IEEE Conference on Computer Vision and
  Pattern Recognition (CVPR)}, June 2018.

\bibitem{edge_2}
Jiaxing Zhao, Jiang-Jiang Liu, Deng-Ping Fan, Yang Cao, Jufeng Yang, and
  Ming-Ming Cheng.
\newblock Egnet: Edge guidance network for salient object detection.
\newblock In {\em Proceedings of the IEEE Conference on Computer Vision and
  Pattern Recognition (CVPR)}, June 2019.

\bibitem{ctd}
Zhirui Zhao, Changqun Xia, Chenxi Xie, and Jia Li.
\newblock Complementary trilateral decoder for fast and accurate salient object
  detection.
\newblock In {\em Proceedings of the 29th acm international conference on
  multimedia}, pages 4967--4975, 2021.

\bibitem{ITSD}
Huajun Zhou, Xiaohua Xie, Jian-Huang Lai, Zixuan Chen, and Lingxiao Yang.
\newblock Interactive two-stream decoder for accurate and fast saliency
  detection.
\newblock In {\em Proceedings of the IEEE Conference on Computer Vision and
  Pattern Recognition (CVPR)}, June 2020.

\bibitem{M_scal_2}
Yuan Zhou, Ailing Mao, Shuwei Huo, Jianjun Lei, and Sun-Yuan Kung.
\newblock Salient object detection via fuzzy theory and object-level
  enhancement.
\newblock {\em IEEE Transactions on Multimedia}, 21(1):74--85, 2019.

\bibitem{BGSAN}
Hongwei Zhu, Peng Li, Haoran Xie, Xuefeng Yan, Dong Liang, Dapeng Chen,
  Mingqiang Wei, and Jing Qin.
\newblock I can find you! boundary-guided separated attention network for
  camouflaged object detection.
\newblock In {\em AAAI Conference on Artificial Intelligence (AAAI)}, February
  2022.

\end{thebibliography}
}

\clearpage
\appendix
Section~\ref{A} shows the experimental details of MFFN on the salient object detection (SOD) task and the test results. We further add more experimental details of MFFN on the COD task and the evaluation metrics changes during training in Section~\ref{D}. 
\section{Experiments on Salient Object Detection}
\label{A}
To illustrate the generalizations and soundness of proposed structural design, we evaluated the proposed model on the salient object detection (SOD) task.

\subsection{Datasets}
Specifically, we only use SOD dataset DUTS-TR~\cite{duts} for MFFN model training, the model performance is evaluated on three existing datasets: ECSSD~\cite{ECSSD}, HKU-IS~\cite{HKUIS}, DUTS-TE~\cite{duts}. The DUTS dataset contains 10553 training images(DUTS-TR) and 5019 test images(DUTS-TE). All the training images are collected from the ImageNet DET training/validation set, while the test images are collected from the ImageNet DET test set and the SUN dataset. The ECSSD dataset contains 1 000 images obtained from the Internet. This dataset is extended by Complex Scene Saliency Dataset (CSSD). Salient objects contain complex structures, and the background has certain complexity. HKU-IS contains 4 447 images, and each image in this dataset meets one of the following three criteria :1) contains multiple scattered salient objects; 2) At least one salient object is in the image boundary; 3) The apparent similarity between the salient objects and the background.

\subsection{Implementation Details}
In order to better evaluate the performance of our model, the compared algorithms are also only trained with DUTS-TR~\cite{duts} and adopt the same hyperparameters and training strategies~\cite{MINet}. Specifically, the initial learning rate is set to 0.05, and follows a linear warm-up and linear decay strategy. In addition, batchsize is set to 8 and trained for 50 epochs, and SGD optimizer is used. 
\subsection{Comparisons with State-of-the-arts}
We compare our proposed model with 10 existing SOD models. As shown in Tab.~\ref{tab:comparesod}, our proposed model outperforms most of the competitors in five evaluation metrics, which indicates that our multi-view strategy can be effectively and efficiently generalized to other segmentation tasks.

\begin{table*}[htbp]
  \centering
  \small
  \caption{Comparison of evaluation results of different Salient object detection(SOD) models on ECSSD~\cite{ECSSD}, HKU-IS~\cite{HKUIS} and DUTS-TE~\cite{duts}. The best results are highlighted in \textcolor{red}{red}, \textcolor{dgreen}{green} and \textcolor{blue}{blue}}
  \resizebox{\linewidth}{!}{
    \begin{tabular}{c|c|ccccc|ccccc|ccccc}
    \toprule
    \multirow{2}[2]{*}{Model} & \multirow{2}[2]{*}{Backbone} & \multicolumn{5}{c}{ECSSD}         & \multicolumn{4}{c}{HKU-IS}    & \multicolumn{6}{c}{DUTS-TE} \\
          &       & $S_m \uparrow$    & $F_\beta^\omega \uparrow$     & $MAE \downarrow$     & $F_\beta \uparrow$     & $E_m \uparrow$     & $S_m \uparrow$     & $F_\beta^\omega \uparrow$     & $MAE \downarrow$     & $F_\beta \uparrow$     & $E_m \uparrow$     & $S_m \uparrow$      & $F_\beta^\omega \uparrow$     & $MAE \downarrow$     & $F_\beta \uparrow$     & $E_m \uparrow$ \\
    \midrule
   
    PAGEnet~\cite{pagenet}&	Vgg16&	0.912& 	0.886& 	0.042& 	0.904& 	0.947& 	0.903& 	0.865& 	0.037& 	0.884& 	0.948& 	0.854& 	0.769& 	0.052& 	0.793& 	0.896 \\
    PiCANet~\cite{picanet}&	ResNet50&	0.917 &	0.867 &	0.046 &	0.890 &	0.952&	0.904 &	0.840 &	0.043 &	0.866 &	0.950&	0.869 &	0.755 &	0.051 	&0.791 &	0.920\\
    PoolNet~\cite{PoolNet}&	ResNet50&	0.926 &	0.904 	&\textcolor{blue}{0.035} &	0.918& 	0.956 &	0.919& 	0.888 &	0.030 &	0.903& 	0.958&	0.887 &	0.817 &	\textcolor{dgreen}{0.037} &	0.840 &	0.926\\
    HRS~\cite{hrs}&	ResNet50&	0.883 	&0.859 	&0.054 	&0.894& 	0.934 &	0.882 &	0.851 &	0.042 &	0.883 &	0.941&	0.829 &	0.746 &	0.051 &	0.791 &	0.899\\
    GCPANet~\cite{gcpanet}&	ResNet50&	\textcolor{blue}{0.927} &	0.903 &	\textcolor{blue}{0.035} &	0.916 &	0.955 &	\textcolor{blue}{0.920}& 	0.889 	&0.031&	0.901 &	0.958&	\textcolor{blue}{0.891} &	0.821 &	\textcolor{blue}{0.038} &	0.841 	&\textcolor{blue}{0.929}\\
    SAMNet~\cite{samnet}&	Handcraft&	0.907 &	0.858 &	0.050 &	0.883 &	0.945 &	0.898 &	0.837 &	0.045 &	0.864 &	0.946&	0.849 &	0.729 &	0.058 &	0.768 &	0.901\\
    VST~\cite{vst}	&T2T-ViTt-14&	\textcolor{red}{0.932} &	0.910 &	{\textcolor{dgreen}{0.033}} &	\textcolor{blue}{0.920} &	\textcolor{red}{0.964} &	\textcolor{red}{0.928} &	0.897 	&\textcolor{blue}{0.029} &	0.907 &	\textcolor{red}{0.968}&	\textcolor{red}{0.896} &	0.828 &	\textcolor{dgreen}{0.037} &	0.845 &	\textcolor{red}{0.939}\\
    Auto-MSFNet~\cite{automsfnet}&	ResNet50&	0.914 &	\textcolor{blue}{0.916} &	{\textcolor{dgreen}{0.033}} &	\textcolor{red}{0.927} &	0.954 &	0.908 &	\textcolor{blue}{0.903} 	&\textcolor{red}{0.027} 	&0.912 	&\textcolor{blue}{0.959}& 	0.877 &	{\textcolor{dgreen}{0.841}} &	\textcolor{red}{0.034} &	\textcolor{blue}{0.855} &	0.931\\
    SGL-KRN~\cite{sgl} &	ResNet50&0.923	& 	0.910 	&	0.036 	&	\textcolor{dgreen}{0.924} 	&	0.954 	&	{\textcolor{dgreen}{0.921}} 	&	{\textcolor{dgreen}{0.904}} 	&	{\textcolor{dgreen}{0.028}}	&	{\textcolor{dgreen}{0.915}} 	&	{\textcolor{dgreen}{0.961}}	&	{\textcolor{dgreen}{0.893}}	&	\textcolor{red}{0.847} 	&	\textcolor{red}{0.034} 	&\textcolor{red}{0.865} 	&	\textcolor{red}{0.939} \\
    CTDNet~\cite{ctd}&	ResNet50&	0.925& 	{\textcolor{dgreen}{0.915}}& 	\textcolor{red}{0.032}& 	\textcolor{red}{0.927}& 	0.956& 	{\textcolor{dgreen}{0.921}}& 	\textcolor{red}{0.909}& 	\textcolor{red}{0.027}& 	\textcolor{red}{0.918}& {\textcolor{dgreen}{	0.961}}&	{\textcolor{dgreen}{0.893}}& 	\textcolor{red}{0.847}& 	\textcolor{red}{0.034}& 	{\textcolor{dgreen}{0.862}}& 	{\textcolor{dgreen}{0.935}} \\
    MINet~\cite{MINet}&	ResNet50&	0.925& 	0.911& 	\textcolor{dgreen}{0.033}& 	0.923& 	\textcolor{blue}{0.957}& 	0.919& 	0.897& 	\textcolor{blue}{0.029}& 	0.909& 0.960&	0.884& 	0.825& 	\textcolor{dgreen}{0.037}& 	0.844& 	0.927 \\
    \textbf{MFFN(ours)}&	ResNet50&	{\textcolor{dgreen}{0.929}}&	\textcolor{red}{0.917}&	\textcolor{red}{0.032}&	\textcolor{red}{0.927}&	{\textcolor{dgreen}{0.959}}&	{\textcolor{dgreen}{0.921}}&	\textcolor{blue}{0.903}&	{\textcolor{dgreen}{0.028}}&	\textcolor{blue}{0.913}&	\textcolor{blue}{0.959}&	0.888&	\textcolor{blue}{0.833}&	\textcolor{blue}{0.038}&	0.850&	0.924\\
    \bottomrule
    \end{tabular}%
    }
  \label{tab:comparesod}%
\end{table*}%

\begin{table*}[t]
  \centering
  \small
  \caption{Our model performs earlystopping at epoch 43, and finally we choose the 40th epoch as our final result, and we provide the following table of the evaluation results between 40th and 43rd epoch.}
  \resizebox{\linewidth}{!}{
    \begin{tabular}{c|ccccc|ccccc|ccccc}
    \toprule
    \multirow{2}[2]{*}{epoch} &  \multicolumn{5}{c}{CHAMELEON}         & \multicolumn{4}{c}{COD10K}    & \multicolumn{6}{c}{NC4K} \\
          & $S_m \uparrow$    & $F_\beta^\omega \uparrow$     & $MAE \downarrow$     & $F_\beta \uparrow$     & $E_m \uparrow$     & $S_m \uparrow$     & $F_\beta^\omega \uparrow$     & $MAE \downarrow$     & $F_\beta \uparrow$     & $E_m \uparrow$     & $S_m \uparrow$      & $F_\beta^\omega \uparrow$     & $MAE \downarrow$     & $F_\beta \uparrow$     & $E_m \uparrow$ \\
    \midrule
   40& 0.905& 	0.852& 	0.021 &	0.871 &	0.963&	0.846 &	0.745 &	0.028 &	0.782 &	0.917&	0.856& 	0.791& 	0.042 &	0.827 &	0.915 \\
    41 & 0.906&	0.850&	0.021	&0.872&	0.965&	0.841&	0.744&	0.030&	0.783&	0.919&	0.854&	0.793&	0.044&	0.825&	0.913 \\
    42 & 0.907&	0.851	&0.023&	0.871&	0.964&	0.844&	0.745&	0.029&	0.782&	0.917&	0.855&	0.790&	0.043&	0.824&	0.916 \\
    43 & 0.907&	0.851&	0.022&	0.870&	0.963&	0.845	&0.746&	0.028&	0.783&	0.918&	0.856&	0.791&	0.042&	0.826&	0.915\\
    \bottomrule
    \end{tabular}%
    }
  \label{tab:earlystop}%
\end{table*}%

\section{Implementation Details and Evaluation Results on COD}
\label{D}
\subsection{Implementation Details}
In this section, we explain the choice of hyperparameters. The weight $\lambda$ of $L_{UAL}$ was initially set as 1.5, and then the cosine strategy is adopted for dynamic adjustment. For comparison with the SOTA model ZoomNet~\cite{ZoomNet-CVPR2022}, we set the initial image size to $384\times384$. The ratio adopted for distance views is 1.5 and 2.0. Finally, the size of FPN output by backbone is ($12\times12$, $24\times24$, $48\times48$, $96\times96$, $192\times192$), and the number of channels is uniformly adjusted to 64. In addition, in the CFU module, the number of interaction groups in \textit{Channel-wise Local Interaction Process} (CLIP) part is 3, and the step of progressive iteration in the \textit{Overall Progressive Iteration} (OPI) is 4. We also encourage readers to experiment with other parameter settings.

\subsection{Early Stopping}

We do not focus on exploiting more epoches as there is no clear evidence that COD detectors will benefit from longer training. During our experiments, we found that the first time for the result dropping was appeared in approximate 40th epoch, as shown in Tab.~\ref{tab:earlystop} and illustrated in Fig.~\ref{fig:my_label}. We also provide the results between 40th epoch and 43rd epoch. To achieve a trade-off between performance and time consumption, we chose the results from the 40th epoch as our final evaluation results. 

\subsection{How to get and evaluat the results of our proposed MFFN?}
\begin{figure*}
    \centering
    \begin{subfigure}[t]{0.46\textwidth}
    \centering
    \includegraphics[width=0.8\textwidth]{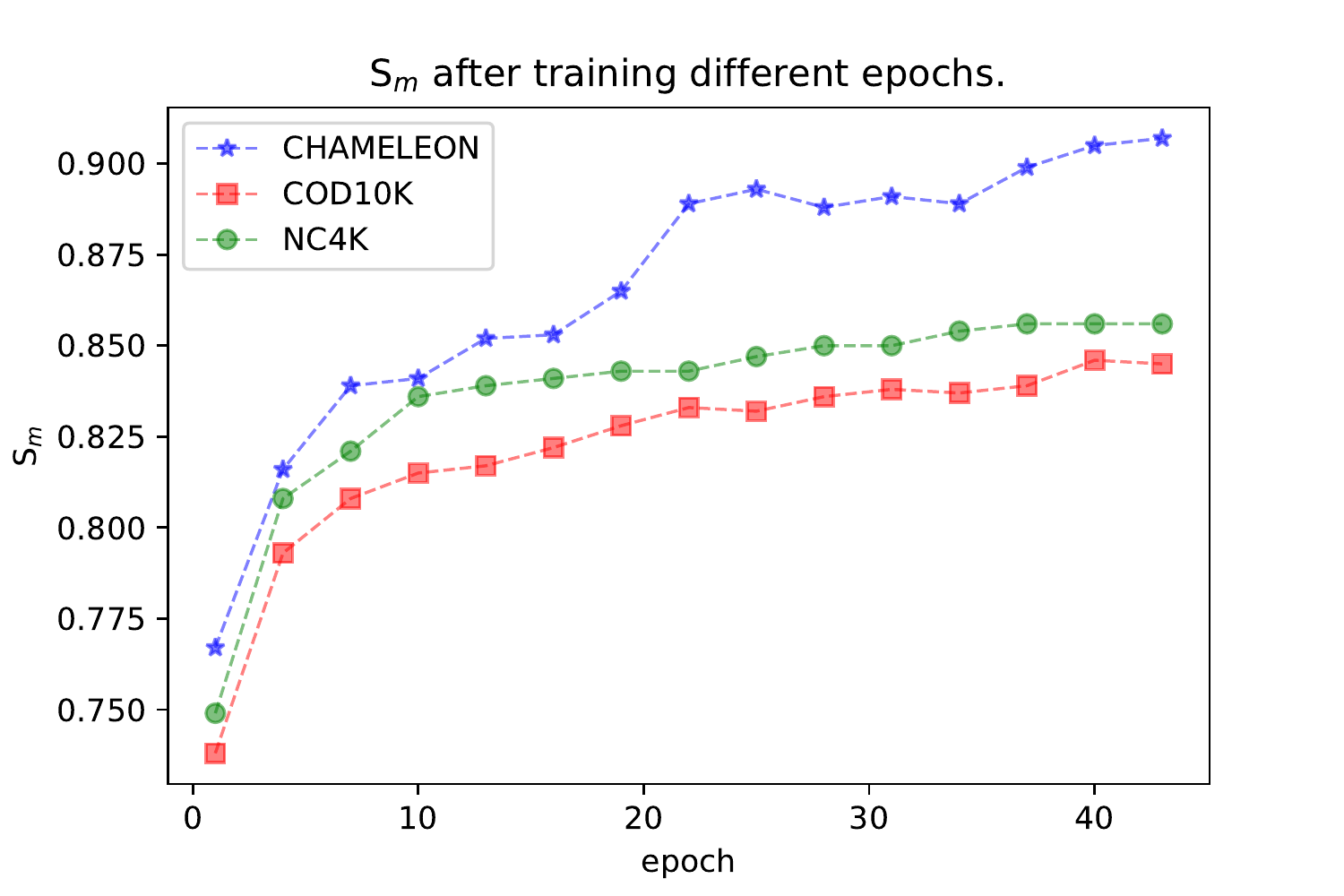} 
    \caption{The change of $F_m$ with the increase of training epochs before the earlystopping.}
    \label{fig:Sm}   
    \end{subfigure}
    \hfill
    \begin{subfigure}[t]{0.46\textwidth}
    \centering\includegraphics[width=0.8\textwidth]{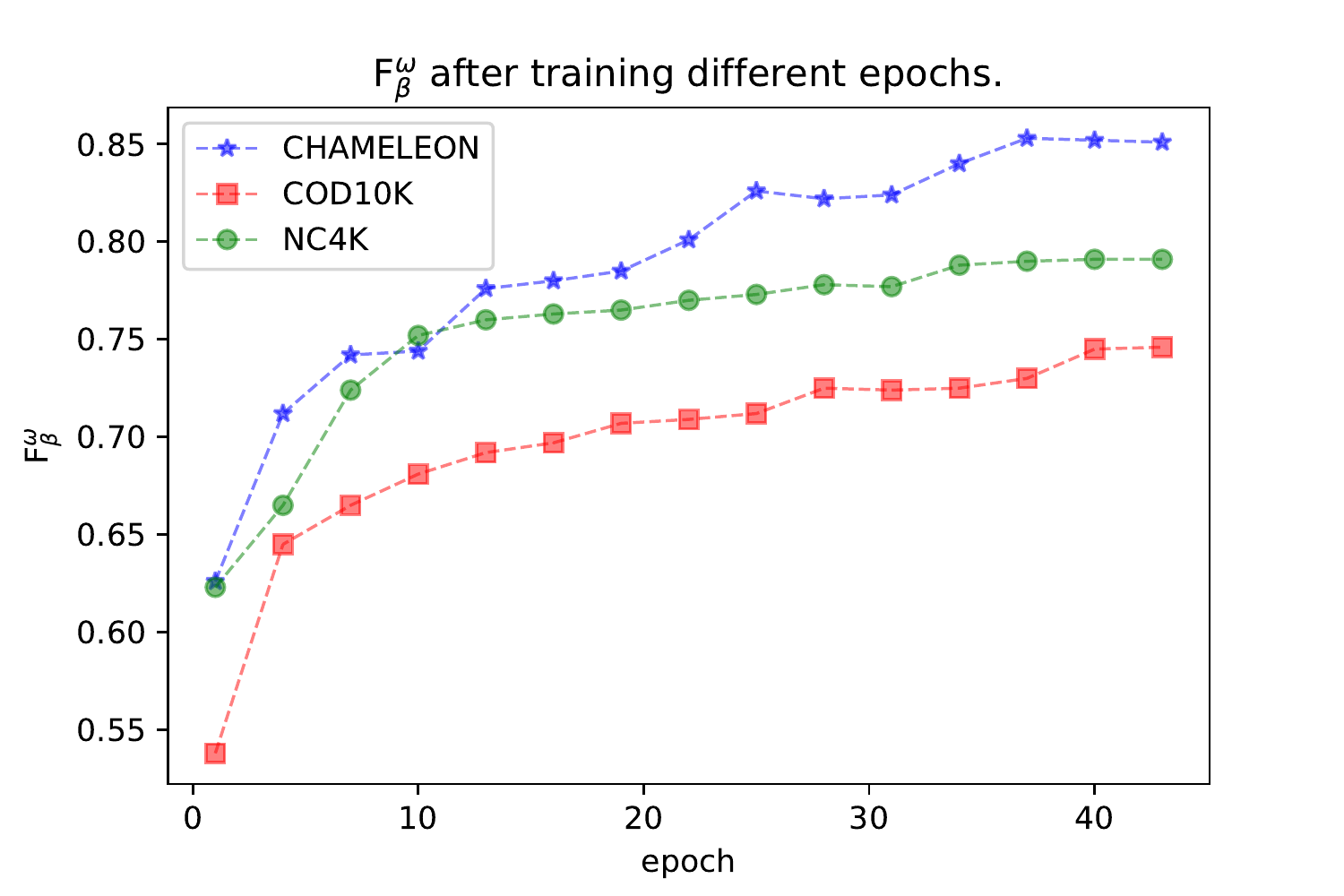} 
    \caption{The change of $F_\beta^\omega$ with the increase of training epochs before the earlystopping.}
    \label{fig:wFb}   
    \end{subfigure}
    \hfill
    \begin{subfigure}[t]{0.46\textwidth}
    \centering\includegraphics[width=0.8\textwidth]{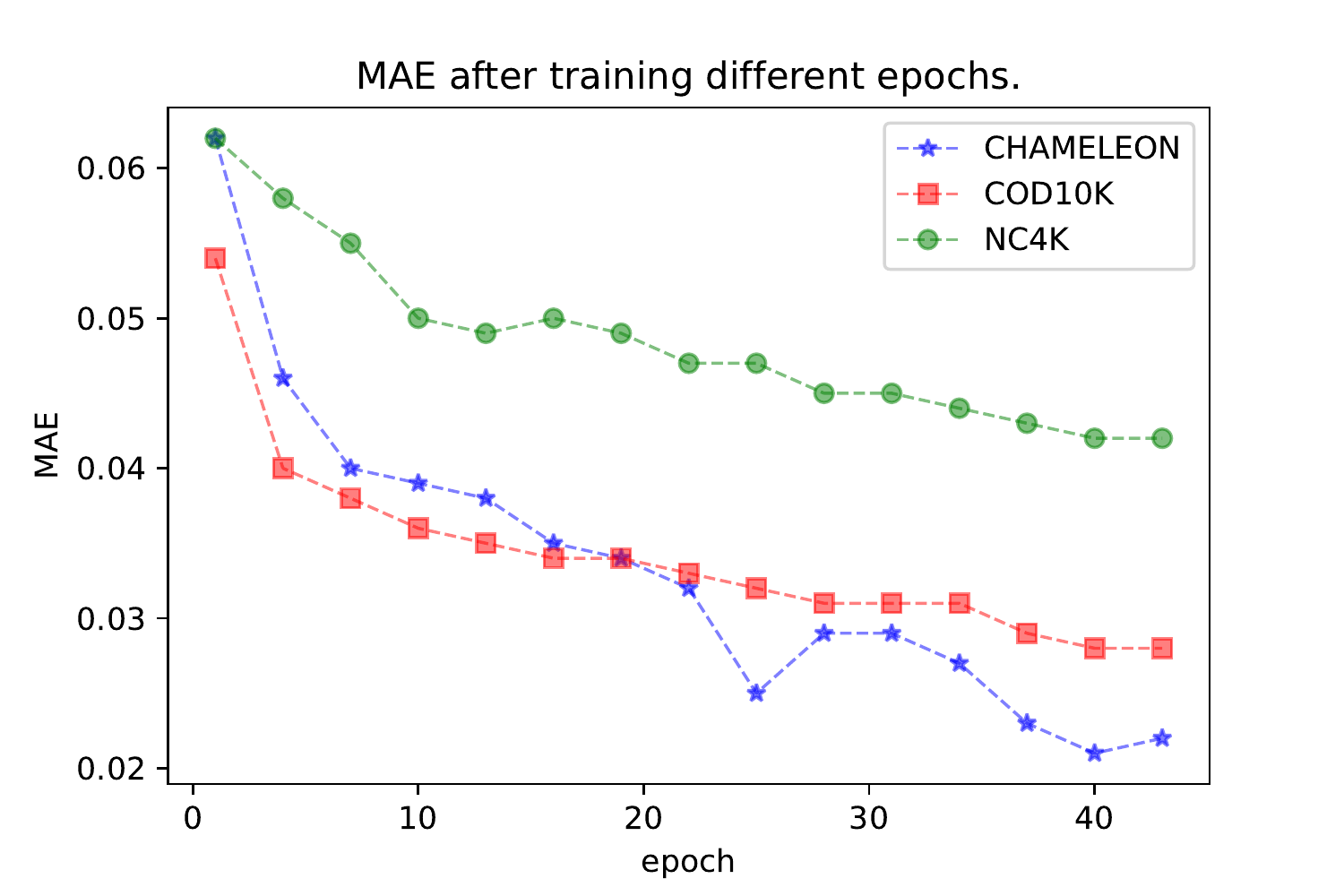} 
    \caption{The change of $MAE$ with the increase of training epochs before the earlystopping.}
    \label{fig:MAE}   
    \end{subfigure}
    \hfill
    \begin{subfigure}[t]{0.46\textwidth}
    \centering\includegraphics[width=0.8\textwidth]{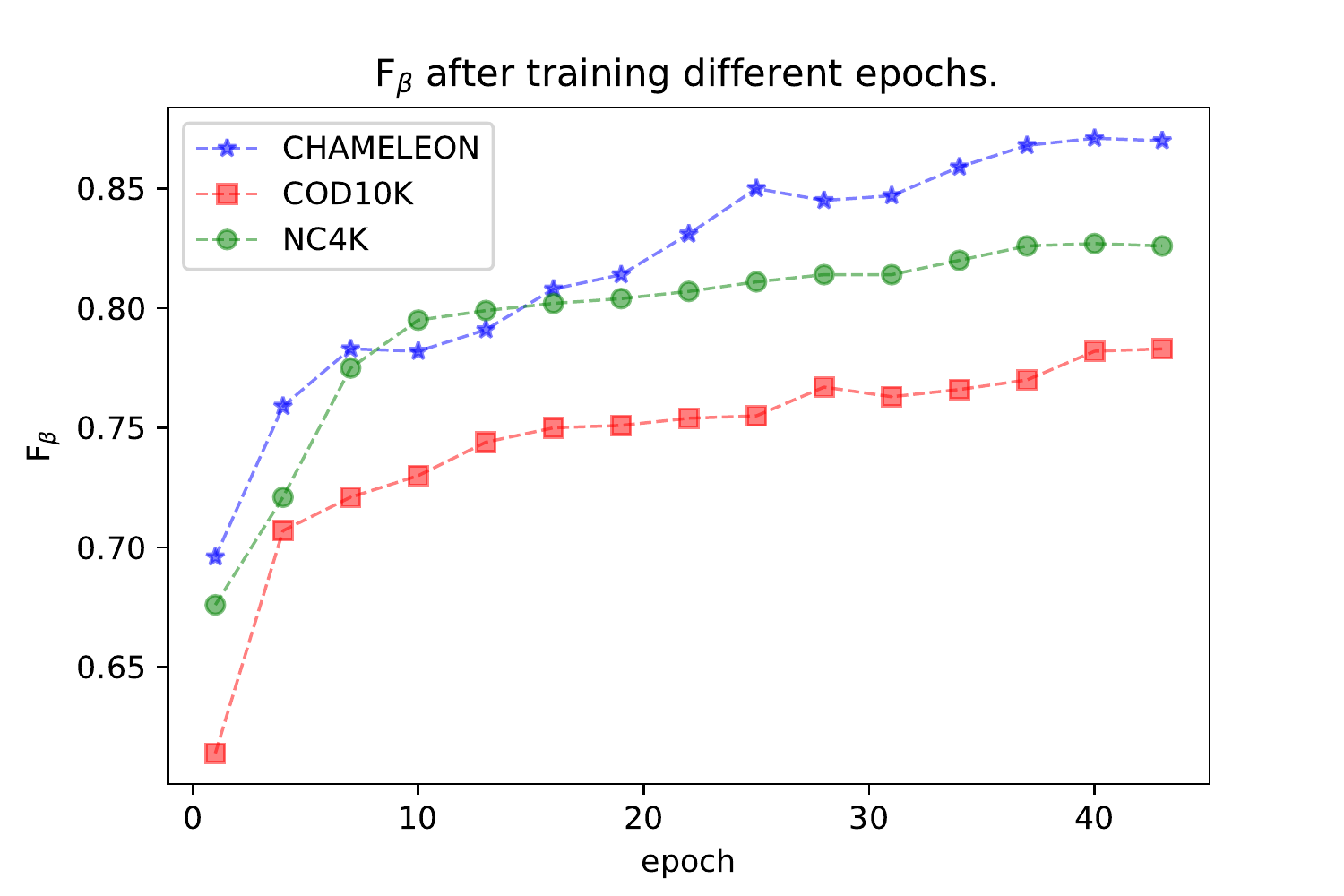} 
    \caption{The change of $F_\beta$ with the increase of training epochs before the earlystopping..}
    \label{fig:Fb}   
    \end{subfigure}
    \hfill
    \begin{subfigure}[t]{0.46\textwidth} 
    \centering\includegraphics[width=0.8\textwidth]{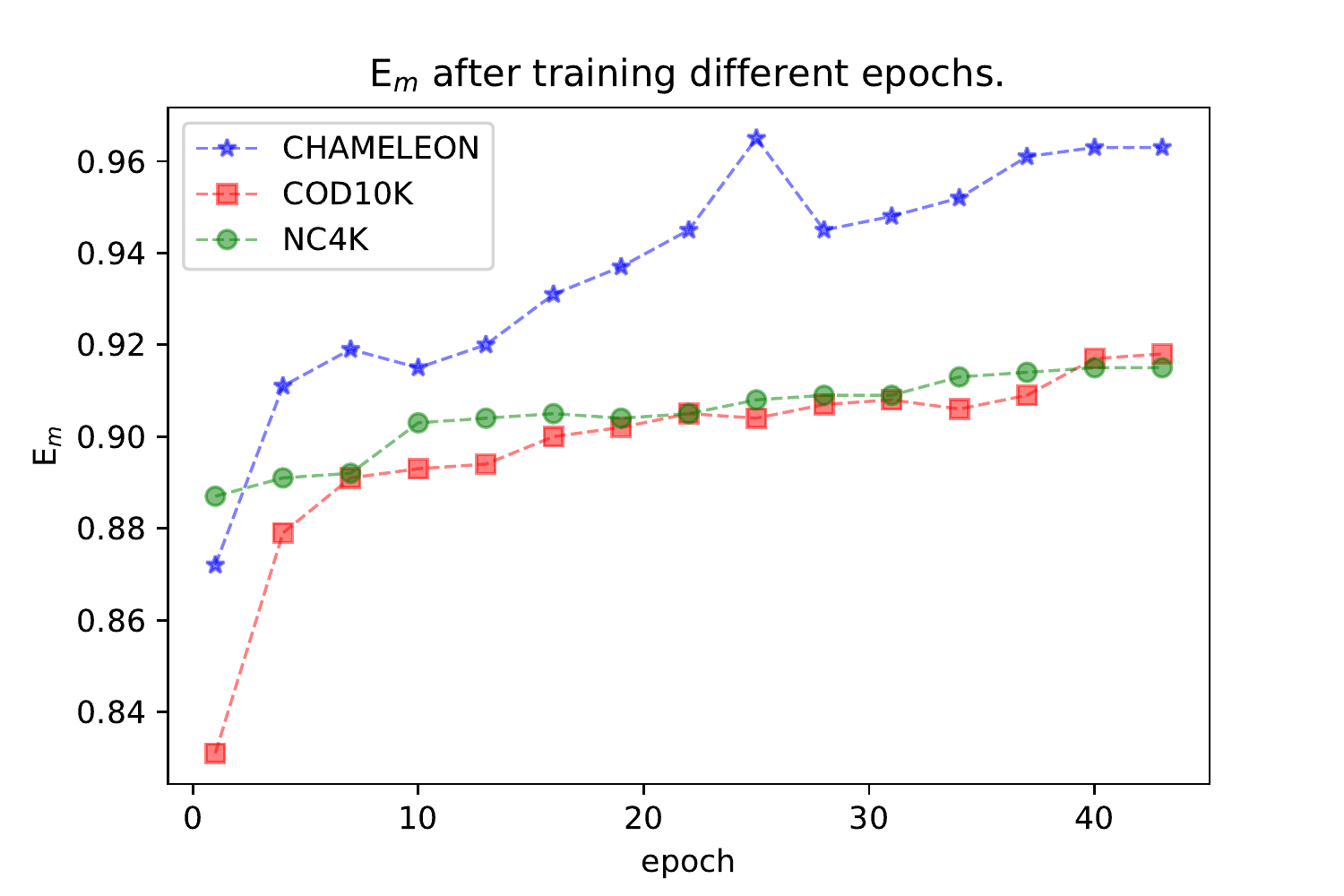} 
    \caption{The change of $E_m$ with the increase of training epochs before the earlystopping.}
    \label{fig:Em}   
    \end{subfigure}
    \caption{Metrics evaluation and early stopping}
    \label{fig:my_label}
\end{figure*}

We use the open source COD evaluation tool to evaluate our prediction results, and we have submitted the test results of the COD10K dataset together with the supplementary material (due to size limitation, we cannot submit the test results of CHAMELEON and NC4K together). You can use the open source tool  \url{https://github.com/DengPingFan/CODToolbox} for evaluation. The changes of five metrics ($F_m$, $F_\beta^\omega$, $MAE$, $F_\beta$, $E_m$) over time (epoch) and \textbf{early stopping} are illustrated in Fig.~\ref{fig:Sm},~\ref{fig:wFb},~\ref{fig:MAE},~\ref{fig:Fb},~\ref{fig:Em}.
\end{document}